\documentclass{article}
\usepackage{neurips}
\usepackage[utf8]{inputenc} 
\usepackage[T1]{fontenc}    
\usepackage[table, dvipsnames]{xcolor}
\definecolor{c1}{HTML}{2F70AF} 
\usepackage[colorlinks=true, citecolor=c1, linkcolor=c1]{hyperref}       
\usepackage{url}            
\usepackage{booktabs}       
\usepackage{microtype}      
\usepackage{amsfonts,amssymb,amsmath,amsthm,bbding}
\usepackage{xspace}

\newcommand{\ie}{\emph{i.e.,}\xspace}

\newcommand{\eg}{\emph{e.g.,}\xspace}

\usepackage{graphicx}
\newcommand{\be}{\mathbf{e}}

\newcommand{\bb}{\mathbf{b}}

\newcommand{\bA}{\mathbf{A}}

\newcommand{\bC}{\mathbf{C}}

\newcommand{\bX}{\mathbf{X}}
\newcommand{\bQ}{\mathbf{Q}}
\newcommand{\bS}{\mathbf{S}}
\newcommand{\bK}{\mathbf{K}}
\newcommand{\bV}{\mathbf{V}}
\newcommand{\bR}{\mathbf{R}}
\newcommand{\sig}{\rlap{$^*$}}
\newcommand{\msig}{\rlap{$^{**}$}}

\newtheorem{definition}{Definition}[section]
\newtheorem{lem}{Lemma}[section]

\usepackage{subfigure}
\usepackage{multirow}
\usepackage{multicol}
\usepackage{caption}
\usepackage{latexsym}
\usepackage{mflogo}
\usepackage{graphics}
\usepackage{enumitem}
\usepackage{algorithm}
\usepackage[noend]{algpseudocode}
\usepackage{wrapfig}
\PassOptionsToPackage{square,numbers,sort&compress}{natbib}

\title{ConvFormer: Revisiting Transformer for \\Sequential User Modeling}
\renewcommand{\shorttitle}{ConvFormer: Revisiting Transformer for Sequential User Modeling}

\usepackage{authblk}

\author[1*]{Hao Wang}
\author[2]{Jianxun Lian}
\author[2]{Mingqi Wu}
\author[3]{Haoxuan Li}
\author[4]{Jiajun Fan}
\author[5]{Wanyue Xu}
\author[2]{Chaozhuo Li}
\author[2]{Xing Xie}
\affil[1]{Zhejiang University}
\affil[2]{Microsoft Research Asia}
\affil[3]{Peking University}
\affil[4]{Tsinghua University}
\affil[5]{Fudan University}
\affil[*]{Corresponding author: \texttt{haohaow@zju.edu.cn}}

\begin{document}

\maketitle

\begin{abstract}

Sequential user modeling, a critical task in personalized recommender systems, focuses on predicting the next item a user would prefer, requiring a deep understanding of user behavior sequences. Despite the remarkable success of Transformer-based models across various domains, their full potential in comprehending user behavior remains untapped. In this paper, we re-examine Transformer-like architectures aiming to advance state-of-the-art performance. We start by revisiting the core building blocks of Transformer-based methods, analyzing the effectiveness of the item-to-item mechanism within the context of sequential user modeling. After conducting a thorough experimental analysis, we identify three essential criteria for devising efficient sequential user models, which we hope will serve as practical guidelines to inspire and shape future designs. Following this,  we introduce ConvFormer, a simple but powerful modification to the Transformer architecture that meets these criteria, yielding state-of-the-art results. Additionally, we present an acceleration technique to minimize the complexity associated with processing extremely long sequences. Experiments on four public datasets showcase ConvFormer's superiority and confirm the validity of our proposed criteria.

\end{abstract}

\section{Introduction}
Recommender system serve as a cornerstone for various online services such as e-commerce~\cite{7927889}, advertising~\cite{zhou2018deep}, and movie \& TV~\cite{10.1145/2843948}. 
In parallel to other tasks as collaborative filtering~\cite{he2017neural} and click-through rate prediction~\cite{xdeepfm}, 
sequential user modeling, which is typically formulated as a next-item-prediction problem, is a foundational task for building effective recommender system. The challenges lie in how to precisely capture evolving preference patterns from user behavior sequence~\cite{sasrec} and then predict what the users will be interested in soon.


Advances in deep learning have spurred the development of sequential user models based on neural networks, represented by recurrent neural networks (RNN)~\cite{gru,repeat}, convolutional neural networks (CNN)~\cite{caser,cnn}, graph neural networks (GNN)~\cite{zhuograph}, and Transformers~\cite{sasrec,bert4rec}. 
Transformer-style models, in particular, have revolutionized fundamental fields with domain-specific adaptations such as the Swin Transformer~\cite{liu2021swin} for images and AlphaFold-v2~\cite{jumper2021highly} for protein structures.
On the contrary, the current progress in sequential user modeling persists in some direct applications of Transformer structure~\cite{bert4rec} with little domain-specific adaptation, and the self-attentive token mixer is surpassed by simpler strategies such as MLP-like modules~\cite{fmlp}.
These observations prompt our investigation into the role of Transformer-like structures, focusing particularly on self-attentive token mixers, within the context of sequential user modeling.

The success of self-attentive token mixers is often ascribed to the scalability and flexibility of the \emph{item-to-item} token mixer paradigm. However, this paradigm's insensitivity to item order (i.e., equivalence to order perturbations) compromises its utility in scenarios where ordering information matters, such as capturing recent and evolving user preferences.
Meanwhile, empirical evidence shows that simple and order-sensitive counterparts (e.g., mixer layers~\cite{mlp4rec} and learnable filters~\cite{gfnet}) produce better performance. Even non-parameterized transformations like FFT~\cite{lee2022fnet} and arbitrary projections~\cite{tay2005synthesizer} offer competitive results. These observations question the indispensability of the item-to-item paradigm and motivate us to mine the core features that support it a prevalent sequential user model.
By deconstructing various aspects of self-attentive token mixer in Section~\ref{sec:empirical}, we identify two factors contributing to its performance: a large receptive field and a lightweight architecture; conversely, the item-to-item paradigm can even impede performance in this context.
On this basis, we propose three criteria for devising token-mixers in sequential user models: order sensitivity, a large receptive field, and a lightweight architecture.

To validate the proposed criteria, we propose \textbf{ConvFormer}, a remarkably simple yet effective modification to the Transformer architecture that satisfies all three criteria.
The core of ConvFormer is replacing the attentive token-mixer with a depth-wise convolution (DWC) layer and enlarging the receptive field. 
Adhering to the proposed criteria, even a straightforward model surpasses various delicate models and achieves state-of-the-art performance, thereby substantiating the efficacy of our criteria.
A potential deficiency would be the computational challenge incurred by the large receptive field. To handle this issue, we develop \textbf{ConvFormer-F}, an efficient approximation based on Fourier transform techniques, which achieves significant speedup with minimal accuracy loss.

To summarize our main contributions:  
\begin{itemize}[leftmargin=*]
    \item We provide a context-specific examination of the self-attentive token mixer and identify three key criteria for designing effective token mixers in sequential user models.
    \item We propose ConvFormer, a simple yet effective update to the vanilla Transformer which is built upon the proposed criteria simultaneously. To address computational challenges, we also propose an acceleration technique based on Fourier transform.
    \item  Through extensive experiments, we demonstrate that ConvFormer outperforms existing models and achieves the state-of-the-art performance. Both the overall performance and ablation studies serve to validate the efficacy of our proposed criteria.
\end{itemize}

\section{Problem statement}
In the context of sequential user modeling, we consider a user $u\in\mathcal{U}$  characterized by a user behavior sequence $S_u=\{i_{1, u},\cdots,i_{\mathrm{L},u}\}$ of length $\mathrm{L}$, which includes chronologically-ordered items with which the user has interacted. 
The objective is to model the likelihood of the next item that the user may interact with, denoted as $p(i_{\mathrm{L}+1}|S_u)$.

We consider the case of item retrieval as recommendations, which models the user representation based on their historical behaviors $S_u$. This user representation serves as a query to retrieve the next item through a simple matching function, such as the dot product.
One exemplar approach to build the user representation is the self-attentive recommender (SAR)~\cite{sasrec}. Let $\mathbf{R}\in\mathbb{R}^{\mathrm{L} \times \mathrm{D}}$ be the embedding of $S_u$ with hidden dimension $\mathrm{D}$, SAR employs a self-attentive token mixer to model the contextual information as follow:
\begin{equation}
    \mathbf{A} = \operatorname{SA}(\mathbf{R}) = \operatorname{Softmax} 
    \left( 
    \left(\mathbf{R} \mathbf{W}^{(\mathrm{Q})}\right)
    \left(\mathbf{R} \mathbf{W}^{(\mathrm{K})}\right)^\top
     / {\sqrt{\mathrm{D}}} 
    \right),
    \label{eq:attention_matrix}
\end{equation}
where $\mathbf{A}\in \mathbb{R}^{\mathrm{L} \times \mathrm{L}}$ is the item-to-item attention matrix. The tokens within the input embedding sequence are then mixed as $\mathbf{S} = \mathbf{A} (\mathbf{R}\mathbf{W}^{(\mathrm{V})})$. $\mathbf{S}$ is further processed through a feed-forward network (FFN), and the building block of SA + FFN can be stacked for multiple layers for deep fusion. The final representation of the last item in $S_u$ serves as the user representation.
\section{Examining Self-attentive token mixer in user modeling}\label{sec:empirical}

\begin{table}
    \centering
	\caption{Performance of SAR and variants. "*" indicates the variants outperforming SAR with p < 0.01. Results on more datasets, configurations and p-values are presented in \autoref{tab:variants}.}\label{tab:itemtoitem}
	\setlength{\tabcolsep}{6mm}{
	\begin{tabular}{llcccccc}
	\toprule
	Dataset                 & Model            & H@5             & H@10             & N@5              & N@10             & MRR              \\ 
        \midrule
\multirow{5}{*}{Sports} 
& SAR   & 0.3442 & 0.4647 & 0.2472 & 0.2861 & 0.2504 \\
& SAR-O  & 0.3474\sig & 0.4682\sig & 0.2497\sig & 0.2887\sig & 0.2526 \\
& SAR-P  & \textbf{0.3478}\sig & \textbf{0.4686}\sig & \textbf{0.2503}\sig & \textbf{0.2891}\sig & \textbf{0.2531}\sig \\
& SAR-R  & 0.3438 & 0.4646 & 0.2470 & 0.2860 & 0.2503 \\
\midrule
\multirow{5}{*}{Yelp} 
& SAR	 & 0.5684 & 0.7446 & 0.4018 & 0.4590 & 0.3841 \\
& SAR-O	 & 0.5713 & 0.7472 & 0.4048 & 0.4618\sig & 0.3870\sig \\
& SAR-P	 & \textbf{0.5731}\sig & \textbf{0.7473} & \textbf{0.4061}\sig & \textbf{0.4626}\sig & \textbf{0.3878}\sig \\
& SAR-R  & 0.5692 & 0.7455 & 0.4033 & 0.4604 & 0.3858 \\

\bottomrule
\end{tabular}
}
\end{table}
This section aims to dissect the self-attentive token mixer to identify its critical components for effective sequential user modeling. Specifically, we focus on three key elements: the item-to-item attentive paradigm, the receptive field, and the overall architecture\footnote{We use the experimental settings in Section~\ref{sec:setup}, \eg setting the embedding size to 64, the maximum sequence length to 50. 
Experiments are repeated 5 times with different seeds.
In the devised toy models, we only modify the token mixer matrix $\mathbf{A}$ in \eqref{eq:attention_matrix} and maintain other tricks identical to vanilla SAR.}.

\subsection{Is the item-to-item token-mixer suitable for user behavior understanding?}
\begin{figure*}
    \centering
    \subfigure[SAR]{\includegraphics[width=0.246\linewidth]{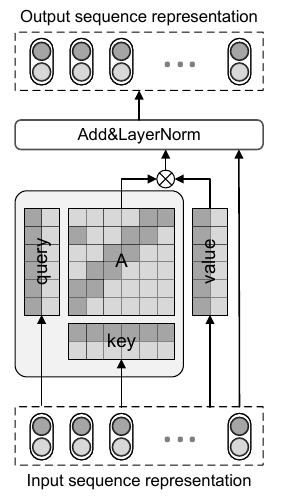}}
    \subfigure[SAR-P]{\includegraphics[width=0.23\linewidth]{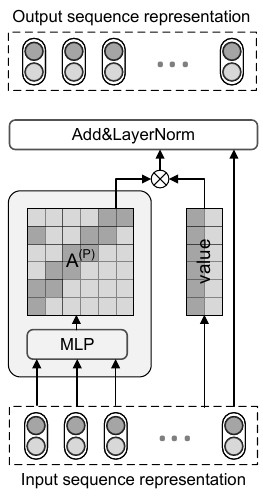}}
    \subfigure[SAR-O]{\includegraphics[width=0.23\linewidth]{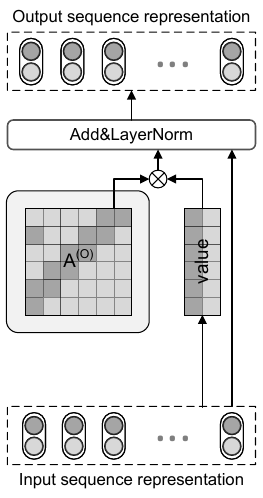}}
    \subfigure[SAR-R]{\includegraphics[width=0.23\linewidth]{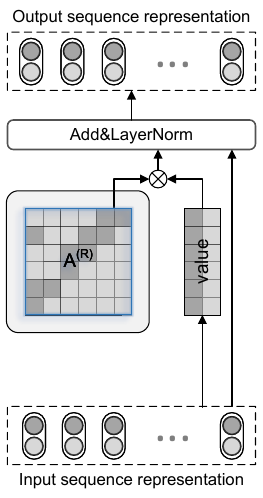}}
    \caption{Simple yet order-sensitive architectures for the alternatives to the item-to-item paradigm in SAR. The non-trainable parameters are indicated by the blue box in (d).}\label{fig:sas}
\end{figure*}
The central element in SAR is the item-to-item attention matrix $\mathbf{A} \in \mathbb{R}^{\mathrm{L} \times \mathrm{L}}$ generated by Eq.(\ref{eq:attention_matrix}). To assess the efficacy of $\mathbf{A}$ for sequential user modeling, we replace it with three alternative mechanisms:
\begin{itemize}[leftmargin=*]
    \item SAR is the standard method~\cite{sasrec} to generate the item-to-item attention matrix. For fair comparison with its alternatives, the multi-head trick is not enabled in this implementation. 
    \item SAR-O (SAR with Order-sensitive weights) utilizes a trainable parameter matrix  $\mathbf{A}^{(\mathrm{O})} \in \mathbb{R}^{\mathrm{L} \times \mathrm{L}}$ which is independent of the input sequence. Unlike SAR which relies solely on position embedding for order information, SAR-O is directly sensitive to the order of the input items. 
    \item SAR-P (SAR with Personalized weights) modifies SAR-O by using MLP to dynamically generate attention scores based on the input $\mathbf{R}$ to this block, wherein $\mathbf{A}^{(\mathrm{P})}[l]=\operatorname{MLP}(\mathbf{R}[l])$. It enables the customization of $\mathbf{A}^{(\mathrm{P})}$ based on inputs, while retaining sensitive to the order of items.
    \item SAR-R (SAR with Random and order-sensitive weights) is similar to SAR-O, but its attention matrix $\mathbf{A}^{(\mathrm{R})} \in \mathbb{R}^{\mathrm{L} \times \mathrm{L}}$ is randomly initialized, kept fixed, and non-trainable. 
\end{itemize}

These models are graphically illustrated in \autoref{fig:sas}. These simpler, order-sensitive alternatives to the item-to-item token mixer in SAR leads to little performance drop as per \autoref{tab:itemtoitem}. 
Notably, SAR-R performs competitively with SAR despite its non-trainable matrix.
Since SAR-R's sole superiority over SAR lies in its order-sensitivity, its competitive performance underscores the importance of order-sensitivity in sequential user modeling.
Furthermore, dynamic weights offer marginal gains (SAR-P vs. SAR-O), suggesting that the adaptive weights in the item-to-item paradigm is not indispensable for SAR's superiority.

Drawing on recent studies that question the necessity of self-attention in various applications~\cite{lee2022fnet,mlpmixer}, our findings suggest that the item-to-item attentive paradigm may constrain the effectiveness of SAR models by neglecting the inherent order of items. Therefore, incorporating architectures that explicitly consider item order has the potential to enhance the performance in sequential user modeling. We provide results on additional datasets and configurations in \autoref{tab:variants}, with statistical test to back up this claim.

\subsection{Is the large receptive field essential for sequential user modeling?}
\begin{figure}[!ht]
    \begin{minipage}{0.49\linewidth}
    \includegraphics[width=\linewidth, trim=15 10 15 10]{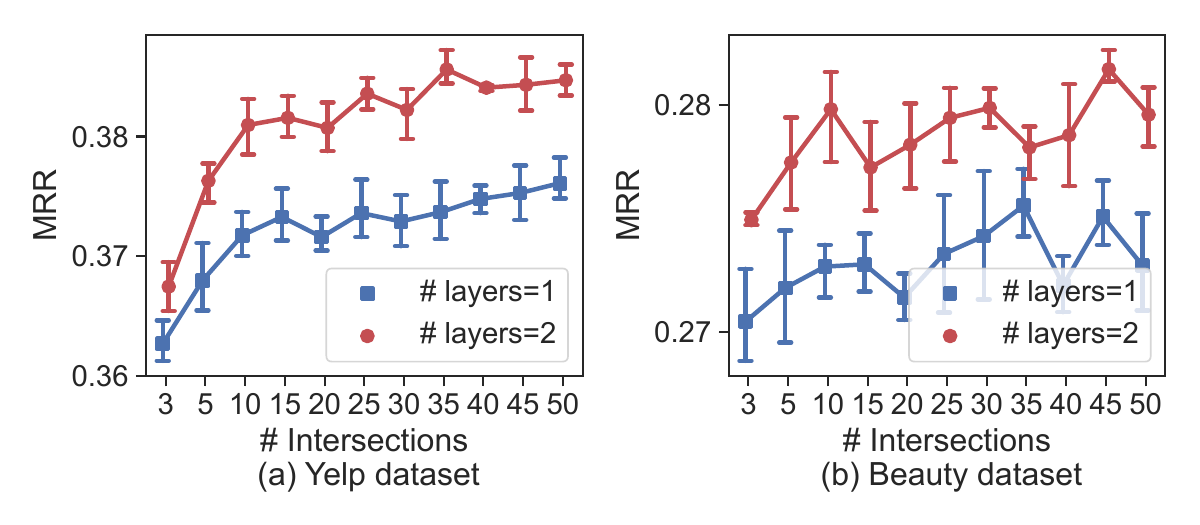}
    \caption{Impact of SAR's receptive field size.
     Error bar denotes 95\% confidence interval.}\label{fig:sas_length}
    \end{minipage}
    \hfill
    \begin{minipage}{0.49\linewidth}
    \includegraphics[width=\linewidth, trim=30 20 30 20]{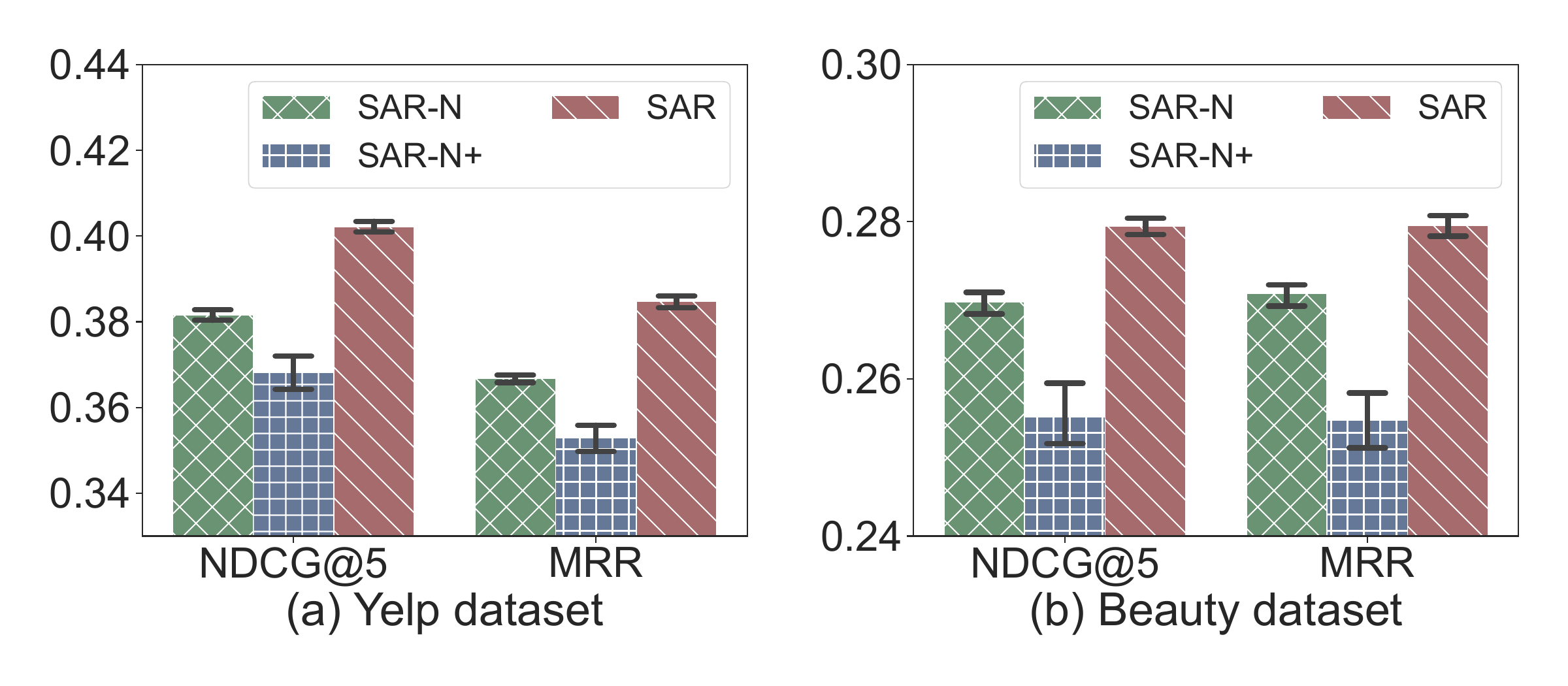}
    \caption{Impact of SAR's lightweight architecture. Error bar denotes 95\% confidence interval.}\label{fig:param_share}
    \end{minipage}
\end{figure}

Another technical distinction of SAR lies on its large receptive field. Each element in the user sequence can directly interact with others within a single self-attention layer, enabling to capture long-term user behavior patterns efficiently. We hypothesize that the large receptive field is a key contributor to SAR's performance. To test this hypothesis, we retain the interactions in SAR's attention matrix between each item and its $\mathrm{K}$ nearest neighbors while deactivating others.
Specifically, we employ a window mask matrix $\Gamma(\mathrm{K})$, wherein $\Gamma_{ij} = 1$ if $|i-j|\leq \mathrm{K}$ and 0 otherwise for $0\leq i, j \leq \mathrm{L}$. Then, replace the attention matrix $\mathbf{A}$ with the element-wise product $\mathbf{A}\odot\Gamma(\mathrm{K})$. 

Our experiments reveal a positive correlation between the size of the receptive field and SAR's performance, as shown in \autoref{fig:sas_length}.
For instance, increasing K from 3 to 45 results in a statistically significant MRR improvement by 4.5\% on the Yelp dataset and 2.43\% on the Beauty dataset. These findings underscore the importance of a large receptive field in enhancing SAR's performance.

\subsection{Is the lightweight architecture essential?} 
Since a large receptive field also increases the model complexity, it is very meaningful to ensure that the overall structure of the model remains lightweight while enlarging the receptive field, to reduce the risk of over-parameterization. We believe that this point is not merely a technical nuance but a pivotal factor in unlocking the advantages of a large receptive field.
For example, SAR exemplifies this by sharing query, key, and value mapping parameters $\textbf{W}^{(*)}$ across all time steps in \eqref{eq:attention_matrix}. 
To empirically validate the importance of a lightweight structure, we introduce two SAR variants:
\begin{itemize}[leftmargin=*]
    \item SAR with non-shared parameters (SAR-N), where the query, key and value mapping parameters (i.e., $\mathbf{W}^{(*)}$ in \eqref{eq:attention_matrix}) are distinct at different time steps.
    \item SAR with more non-shared parameters (SAR-N+),  where all items in the input sequence are concatenated to generate query, key, and value vectors.
\end{itemize}   
Both variants above sacrifice the lightweight of vanilla SAR for order-sensitivity and high capacity. However, according to \autoref{fig:param_share}, they exhibit a decline compared to the original SAR model. Specifically, SAR-N+ shows a relative MRR decrease of 3.11\% on Yelp and 8.91\% on Beauty, while SAR-N experiences a relative MRR drop of 4.65\% on Yelp and 8.27\% on Beauty. 
These findings underscore the necessity of a lightweight architecture in mitigating the risks associated with large receptive fields, thereby preserving the efficacy of SAR models.

\section{Proposed method}
\subsection{Three criteria for sequential user modeling}\label{sec:crit}
\begin{wrapfigure}{r}{6.5cm}
    \centering
    \vspace{-0.5cm}
	\includegraphics[width=1\linewidth]{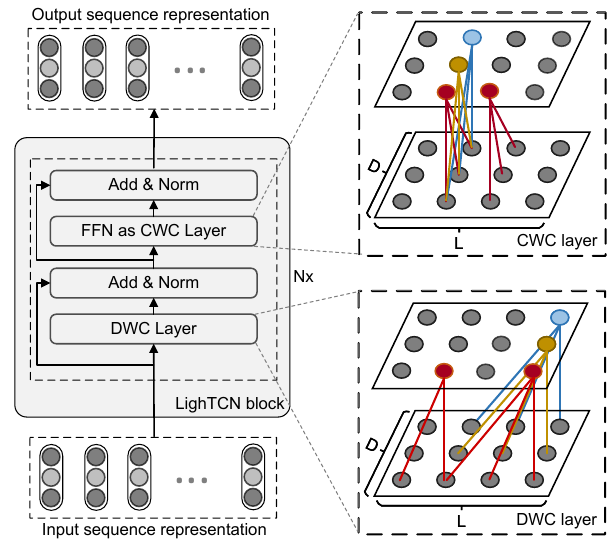}
    \caption{The core structure of ConvFormer.}
    \vspace{-0.8cm}
    \label{fig:model-overview}
\end{wrapfigure}
The empirical studies in Section~\ref{sec:empirical} suggest that a large perception field and a lightweight architecture are key factors that contribute to the superior performance of SAR. However, the item-to-item paradigm is identified as a limitation due to its insensitivity to item order. In light of these findings, we summarize three principles for designing effective token mixers in sequential user modeling:
\begin{enumerate}[leftmargin=*]
   \item The token-mixer should be sensitive to the order of items, to capture the sequential patterns such as evolving and transient preference from user behaviors;
    \item The token-mixer should have a large receptive field, to capture and exploit the long-term patterns in user behavior sequences;
    \item The token-mixer should maintain a lightweight architecture, to mitigate the risk of overfitting resulting from a large receptive field.
\end{enumerate}

Prevalent sequential user models fail to meet the criteria simultaneously: SAR-based and RNN-based solutions fail to meet criteria (1) and (2), respectively; CNN-based methods like Caser~\cite{caser} employ narrow receptive field and conventional convolution operators, failing to meet criteria (2) and (3).
Given these observations, we find that current architectures have significant room for improvement in aligning with the proposed criteria. 

\subsection{The ConvFormer architecture}
In light of these, we develop ConvFormer, a simple yet effective update to the vanilla SAR architecture. 
The primary technical contribution is the replacement of the item-to-item mechanism with a novel LighTCN layer, which involves a large receptive field and order-sensitive item interactions while remaining lightweight overall architecture, meeting the proposed criteria simultaneously. \autoref{fig:model-overview} presents the core building block of ConvFormer, and the workflow is described below.


\subsubsection{Embedding layer}

We maintain an item embedding look-up table $\mathbf{E}^{(\mathrm{I})} \in \mathbb{R}^{\mathcal{I} \times \mathrm{D}}$ to map the high-dimensional, one-hot item indices to a low-dimensional, dense representation space.
We also add a learnable position encoding matrix $\mathbf{E}^{(\mathrm{P})} \in \mathbb{R}^{\mathrm{L} \times \mathrm{D}}$ to incorporate the ordering information into the item semantics.
Formally, a user's historical behaviors can be represented as 
\begin{equation}\label{eq:embed1}
    \hat{\mathbf{E}}=[\mathbf{E}^{(\mathrm{I})}_{i_1}+\mathbf{E}^{(\mathrm{P})}_1, \mathbf{E}^{(\mathrm{I})}_{i_2}+\mathbf{E}^{(\mathrm{\mathrm{P}})}_2, ..., \mathbf{E}^{(\mathrm{I})}_{i_\mathrm{L}}+\mathbf{E}^{(\mathrm{P})}_\mathrm{L}],
\end{equation}
where $\mathbf{E}^{(\mathrm{I})}_{i_j}$ is the item embedding of the $j$-th item in $S_{u}$. If the sequence length is less than $\mathrm{L}$, pad zeros on the left side.
To avoid overfitting and ensure a stable training process, we follow the common setting~\cite{fmlp} to refine (\ref{eq:embed1}) with dropout and layer normalization:
\begin{equation}\label{eq:embed2}
    \hat{\mathbf{E}}=\operatorname{Dropout}(\operatorname{LayerNorm}(\mathbf{E}^{(\mathrm{I})}+\mathbf{E}^{(\mathrm{P})})).
\end{equation}

\subsubsection{Light temporal convolution neural (LighTCN) layer}

Following the embedding layer, we extract the user embedding by stacking multiple LighTCN layers. 
A LighTCN layer consists of two sub-layers, \ie a depth-wise convolution layer and a channel-wise convolution layer. 
In the depth-wise convolution (DWC) layer, we perform convolution operation for each channel\footnote{We view each dimension of the embedding vector as a single channel, using the term "channel" to align with terminology in the computer vision domain.} individually.
Specifically, let $\bR\in \mathbb{R}^{\mathrm{L}\times \mathrm{D}}$ be the sequence of input representation for a certain layer (for the first layer, $\bR = \hat{\mathbf{E}}$), $\bC\in \mathbb{R}^{\mathrm{K}\times \mathrm{D}}$ be the convolution kernel with size $\mathrm{K}$, we first conduct the depth-wise convolution operator $\operatorname{DWC}(\cdot)$ along the temporal axis as follow:
\begin{equation}\label{eq:dwc-vani}
    \operatorname{DWC}(\textbf{R})_{l,d} = \operatorname{Pad}(\sum_{k=1}^\mathrm{K}{\bR_{l+k-1,d} * \bC_{k,d}}), \quad d=1,\dots,\mathrm{D},
\end{equation}
where the output is left padded to ensure $\operatorname{DWC}(\bR)\in\mathbb{R}^{\mathrm{L}\times \mathrm{D}}$. 
Following the meta-former paradigm~\cite{yu2022metaformer}, the skip connection, layer normalization and dropout are incorporated:  
\begin{equation}\label{eq:dwc01}
    \hat{\bR} = \operatorname{LayerNorm}(\bR + \operatorname{Dropout}(\operatorname{DWC}(\bR)).
\end{equation}

The DWC operator extracts the linear temporal characteristics on each individual channel, while overlooking the non-linearity and channel-wise intersections. Thus, we employ the FFN layer in Transformers~\cite{vaswani2017attention}  to model the non-linear channel-wise intersections. Notably, FFN can be implemented with a channel-wise convolution (CWC) operation, consisting of a $1\times1$ convolution operator $f(\mathbf{x})=\mathbf{x}\mathbf{W}+\mathbf{b}$ and a ReLU activation function. For consistency with DWC, we denote FFN as CWC in the LighTCN block.
At each time step, the CWC layer is computed as
\begin{equation}
    \operatorname{CWC}(\hat{\bR}_l) = \operatorname{FFN}(\hat{\bR}_l) = f(\operatorname{ReLU}(f(\hat{\bR_{l}})), \quad l=1,2,\dots, \mathrm{L},
\end{equation}
which is similarly refined by the skip connection, dropout and layer normalization technologies:
\begin{equation}\label{eq:dwc}
    \tilde{\bR} = \operatorname{LayerNorm}(\hat{\bR} + \operatorname{Dropout}(\operatorname{CWC}(\hat{\bR})).
\end{equation}

\subsubsection{Dot-product scorer}
LighTCN, as a building block, can be stacked for learning more complicated interactions, as shown in \autoref{fig:model-overview}. Finally, let $\be_{c}$ be the embedding of the item $c$ in $\mathbf{E}^{(\mathrm{I})}$, $\bar{\bR}$ be the output of the last LighTCN layer. Following the two-tower retrieval paradigm~\cite{sasrec,fmlp}, we use $\bar{\bR}[\mathrm{L}]$, the output at the last step, as the user representation. We estimate the likelihood of user interacting with item $c$ at the $\mathrm{L}+1$ step as $p(i_{\mathrm{L}+1}=c|i_{1:\mathrm{L}})=\be_{c}^{\top}\bar{\bR}[\mathrm{L}]$.
Finally, we update learnable weights to minimize the pairwise ranking loss at a generative paradigm:
\begin{align}
\label{ft}
    \mathcal{L} = -\sum_{u \in \mathcal{U}}\sum_{l=1}^{\mathrm{L}}\log \sigma\bigg(p(i_{l+1}|i_{1:l})-p(i_{l+1}^{-}|i_{1:l})\bigg),
\end{align}
where each ground-truth item $i_{l+1}$ is paired with a negative item $i_{l+1}^{-}$ that is randomly sampled.

\subsection{Accelerated approximation algorithm}\label{sec:acceleration}

One potential concern of the LighTCN layer, regarding the criterion (2), is the computational cost of large receptive field given lengthy behavior sequences, as highlighted by recent studies~\cite{pi2020search,chen2021end}. This is because as the receptive field broadens to almost the sequence length, the computational complexity of the model becomes $\mathcal{O}(\mathrm{L}^2)$. To address this issue, we have developed an acceleration algorithm for ConvFormer, denoted by \textbf{ConvFormer-F}, inspired by the convolution theorem~\cite{fcnn,dsp} in Lemma~\ref{lem:a1}.
The key inspiration of Lemma~\ref{lem:a1} is that the convolution in the temporal domain can be transformed into a Hadamard product in the Fourier domain, leading to a more efficient computation of the DWC layer:
\begin{equation}\label{eq:accelerate}
    \operatorname{DWC}(\mathbf{R}) = \mathcal{F}^{-1}\left(\mathcal{F}(\bR) \odot \mathcal{F}(\bC)\right)
\end{equation}
where $\odot$ indicates the Hadamard point-wise product,  $\mathbf{C}$ is right-padded with zeros to ensure it has the same length as $\mathbf{R}$. $\mathcal{F}$ denotes the Discrete Fourier Transform~\cite{dsp}, a fundamental technique for processing discrete time series data~\cite{dsp}, and $\mathcal{F}^{-1}$ indicates the inverse DFT (IDFT) in Definition~\ref{def:dft}. 

\begin{definition}[DFT and IDFT]\label{def:dft}
Given an $\mathrm{L}$-length sequence $\mathbf{X}=\left[x_1,...,x_\mathrm{L}\right]$, DFT projects it to a set of predefined exponential basis, and the projection onto the $k$-th basis is calculated as
\begin{equation}
    \label{eq:dft}
    x_k^{(\mathrm{F})} = \mathcal{F}(\mathbf{X})_{k} = \sum_{l=0}^{\mathrm{L}-1} x_l \exp(-\frac{2\pi i}{\mathrm{L}} l k), \quad 0 \leq k \leq \mathrm{L}-1,
\end{equation}
where $\exp(\cdot)$ is the exponential basis, $i$ is the imaginary unit, $k$ indicates the frequency of the exponential basis.
Inversely, given the projection onto each basis, we can recover the original sequence via the Inverse DFT (IDFT):
\begin{equation}
    x_l = \mathcal{F}^{-1}(\mathbf{X}^{(\mathrm{F})})_{l} = \frac{1}{\mathrm{L}} \sum_{k=0}^{\mathrm{L} - 1}x_k^{(\mathrm{F})} \exp({\frac{2\pi i}{\mathrm{L}} lk}), \quad 0 \leq l \leq \mathrm{L}-1,
    \label{equ:idft}
\end{equation}
\end{definition}

Both DFT and IDFT can be implemented as matrix-vector multiplication, which is differentiable and thus can be integrated in neural user models. However, the complexity of DFT and IDFT is $\mathcal{O}(\mathrm{L}^2)$, with no theoretical superiority over the standard DWC layer.
In this regard, the actual accelerator are the Fast Fourier Transform (FFT) and its inverse, which calculate DFT and IDFT in a recursive manner and reduce their complexity
to $\mathcal{O}(\mathrm{L}\log(\mathrm{L}))$.
In this way, we can reduce the computational complexity of the DWC layer from $\mathcal{O}(\mathrm{L}^2)$ to $\mathcal{O}(\mathrm{L}\log(\mathrm{L}))$, which is extremely advantageous when modelling long user behavior sequences. 

The computational workflow of ConvFormer and ConvFormer-F is summarized in Algorithm~\ref{alg:convformer}. 
Overall, ConvFormer-F can greatly improve the efficiency of ConvFormer in handling lengthy user behavior sequences while maintaining accuracy. We verify this claim in Section~\ref{sec:speed}.

\begin{algorithm}[tb]
\flushleft
\caption{The computational workflow of ConvFormer}\label{alg:accelerate}
\label{alg:convformer}
\textbf{Input}: a user's sequence $S=\{i_{1},\cdots,i_{\mathrm{L}}\}$, a target item $i_t$.\\
\textbf{Output}: the preference score $p(i_t|i_{1:\mathrm{L}})$.
\begin{algorithmic}[1] 
\State get input embeddings $\hat{\mathbf{E}}$ of $S$ by Eq.(\ref{eq:embed2})
\State set $\bR \leftarrow \hat{\mathbf{E}}$ and lookup target item's embedding $\be_{t}$  
\For {$n=1$ to $\mathrm{N}$} \Comment{Stacking $\mathrm{N}$ LighTCN layers}
\If{acceleration}\Comment{Based on Eq.(\ref{eq:accelerate})}
\State $\mathbf{C}^{(\mathrm{F})}\leftarrow \mathcal{F}(\operatorname{Pad}(\mathbf{C})), \quad \mathbf{R}^{(\mathrm{F})}\leftarrow \mathcal{F}(\mathbf{R})$
\State $\hat{\bR} \leftarrow \operatorname{LayerNorm}(\bR + \operatorname{Dropout}(\mathcal{F}^{-1}(\mathbf{C}^{(\mathrm{F})}\odot \mathbf{R}^{(\mathrm{F})}))$
\Else\Comment{Standard operation with Eq.(\ref{eq:dwc-vani})}
\State $\hat{\bR} \leftarrow \operatorname{LayerNorm}(\bR + \operatorname{Dropout}(\operatorname{DWC}(\mathbf{R}))$
\EndIf
\State $\tilde{\bR} \leftarrow \operatorname{LayerNorm}(\hat{\bR} + \operatorname{Dropout}(\operatorname{CWC}(\hat{\bR}))$
\State $\bR \leftarrow \tilde{\bR}$
\EndFor
\State $p(i_t|i_{1:\mathrm{L}})=\be_{t}^{\top}\bR[L]$ \Comment{Dot product scorer}
\end{algorithmic}
\end{algorithm}
\section{Experiments}
To demonstrate the efficacy of both the proposed criteria and ConvFormer, which is a simple yet inspiring model built upon these criteria, the five aspects as follows deserve empirical investigation. 
\begin{itemize}[leftmargin=*]
    \item \textbf{Performance:} \textit{Does ConvFormer work?} We compare ConvFormer against state-of-the-art baselines, with the 1-vs-99 performance in \autoref{tab:main_table} and the full-sort performance in Section~\ref{sec:full}.
    \item \textbf{Gains:} \textit{Why does it work?} We deconstruct various aspects of ConvFormer in Section~\ref{sec:ablation} to identify the sources of its accuracy gain and back up the efficacy of the proposed three criteria. We provide additional comparisons and rigorous statistical tests in Appendix \ref{sec:sarvariant}.
    \item \textbf{Generality:} \textit{Does it work in other datasets and tasks?} We investigate the performance on real-world Xbox recommendation scenario in Section \ref{sec:industrial}, and a general CTR prediction task in Appendix \ref{sec:ctr}.
    \item \textbf{Speed:} \textit{Does ConvFormer-F reduces running time while preserving accuracy?} We compare actual running time of SASRec, ConvFormer and ConvFormer-F in various settings in Appendix \ref{sec:speed}.
    
\end{itemize}

\subsection{Experimental setup}\label{sec:setup}

\paragraph{Dataset.}
We perform experiments on four sequential user modeling datasets that cover a diverse range of domains. \textbf{ Beauty, Sports, and Toys} are three datasets extracted from the Amazon review data source~\cite{DBLP:conf/sigir/McAuleyTSH15}. The \textbf{Yelp} is a public dataset which consists of user interactions with local businesses, such as restaurants, bars and cafes, in the form of user reviews. 
Each user's interactions are organized chronologically, with the latest item set aside for testing, the second to last item designated for validation, and the remaining items used for training. Users or items with less than five interactions are excluded. The processed datasets' statistics are summarized in \autoref{tab:datasets}.

\begin{table}
\centering
\caption{Statistics of the employed datasets.}
\label{tab:datasets}
    \setlength{\tabcolsep}{5mm}{
    \small
    \begin{tabular}{lllll}
    \toprule
    \textbf{Dataset} & \#.\textbf{Sequences} & \#.\textbf{Items} & \#.\textbf{Actions} & \textbf{Sparsity} \\
    \midrule
    Beauty & 22,363 & 12,101 & 198,502 & 99.93\% \\
    Sports & 25,598 & 18,357 & 296,337 & 99.95\% \\
    Toys & 19,412 & 11,924 & 167,597 & 99.93\% \\
    Yelp & 30,431 & 20,033 & 316,354 & 99.95\% \\
    Xbox &674,491&9,690 & 19,699,497 & 99.70\% \\
\bottomrule
\end{tabular}}
\end{table}

\paragraph{Evaluation Protocol.}
We employ three ranking metrics for evaluation: Top-$k$ hit ratio (H@$k$), Top-$k$ normalized discounted cumulative gain (N@$k$), and mean reciprocal rank (MRR). 
As for the negative item candidates, we experiment on both types of settings: (1) ranking the positive item against 99 randomly selected non-interacted items for each user; and (2) the full-sort test set where the positive item is ranked alongside all non-interacted items.

\paragraph{Baseline Models.}
The collection of baselines includes\footnote{We respect existing benchmark results, following the settings, datasets and baselines \cite{fmlp}.}: 
(1) \textbf{PopRec},  \textbf{FM}~\cite{fm}, and \textbf{AutoInt}~\cite{autoint} are non-sequential models;
(2) \textbf{GRU4Rec}~\cite{gru}, \textbf{Caser}~\cite{caser}, \textbf{HGN}~\cite{hgn}, 
\textbf{CLEA}~\cite{clea}, and \textbf{SRGNN}~\cite{srgnn} are representative sequential baselines which do not involve Transformer architectures;
(3) \textbf{SASRec}~\cite{sasrec}, \textbf{BERT4Rec}~\cite{bert4rec}, \textbf{GCSAN}~\cite{gcsan} and \textbf{FMLP-Rec}~\cite{fmlp} are baselines that (partially) rely on Transformer architectures.

\subsection{Overall performance}\label{sec:perform}
\subsubsection{1-vs-99 test test on public datasets}

\begin{table*}[t!]
    \scriptsize
    \centering
	\caption{Performance comparison on four datasets. The bold and underlined fonts indicate the best and second-best performance, respectively.
    “*” and "**" mark the metrics where ConvFormer outperforms the best baselines with p-value $<$ 0.05 and 0.001, respectively, in the one-sample t-test.}
	\label{tab:main_table}
	\setlength{\tabcolsep}{1mm}{
    
	\begin{tabular}{llcccccccccccccc}
	\toprule
    Dataset & Metric & PopRec &FM& AutoInt &GRU4Rec & Caser &HGN &CLEA &SASRec &BERT4Rec &SRGNN &GCSAN &FMLP-Rec &ConvFormer  \\
	\midrule
\multirow{6} * {Beauty}
 &H@1   &0.0678 &0.0405 &0.0447 &0.1337 &0.1337  &0.1683 &0.1325 &0.1870 &0.1531 &0.1729 &{0.1973} &\underline{0.2011} &\textbf{0.2019}\\
 &H@5   &0.2105 &0.1461 &0.1705 &0.3125 &0.3032  &0.3544 &0.3305 &0.3741 &0.3640 &0.3518 &0.3678 &\underline{0.4025}  &{\textbf{0.4119}}\msig\\
 &N@5 &0.1391 &0.0934 &0.1063 &0.2268 &0.2219  &0.2656 &0.2353 &0.2848 &0.2622 &0.2660 &0.2864 &\underline{0.3070}  &{\textbf{0.3125}}\msig\\
 &H@10  &0.3386 &0.2311 &0.2872 &0.4106 &0.3942  &0.4503 &0.4426 &0.4696 &0.4739 &0.4484 &0.4542 &\underline{0.4998}  &{\textbf{0.5105}}\msig\\
 &N@10&0.1803 &0.1207 &0.1440 &0.2584 &0.2512  &0.2965 &0.2715 &0.3156 &0.2975 &0.2971 &0.3143 &\underline{0.3385}  &{\textbf{0.3443}}\msig\\
 &MRR    &0.1558 &0.1096 &0.1226 &0.2308 &0.2263  &0.2669 &0.2376 &0.2852 &0.2614 &0.2686 &0.2882 &\underline{0.3051}  &{\textbf{0.3093}}\sig\\
\midrule
\multirow{6} * {Sports}
 &H@1   &0.0763 &0.0489 &0.0644 &0.1160 &0.1135  &0.1428 &0.1114 &0.1455 &0.1255 &0.1419 &\underline{0.1669} &{0.1646} &\textbf{0.1671}\\
 &H@5   &0.2293 &0.1603 &0.1982 &0.3055 &0.2866  &0.3349 &0.3041 &0.3466 &0.3375 &0.3367 &{0.3588} &\underline{0.3803} &\textbf{0.3891}\msig\\
 &N@5 &0.1538 &0.1048 &0.1316 &0.2126 &0.2020  &0.2420 &0.2096 &0.2497 &0.2341 &0.2418 &{0.2658} &\underline{0.2760} &\textbf{0.2819}\msig\\
 &H@10  &0.3423 &0.2491 &0.2967 &0.4299 &0.4014  &0.4551 &0.4274 &0.4622 &0.4722 &0.4545 &{0.4737} &\underline{0.5059} &\textbf{0.5116}\msig\\
 &N@10  &0.1902 &0.1334 &0.1633 &0.2527 &0.2390  &0.2806 &0.2493 &0.2869 &0.2775 &0.2799 &{0.3029} &\underline{0.3165} &\textbf{0.3215}\msig\\
 &MRR    &0.1660 &0.1202 &0.1435 &0.2191 &0.2100 &0.2469 &0.2156 &0.2520 &0.2378 &0.2461 &{0.2691} &\underline{0.2763} &\textbf{0.2808}\msig\\
\midrule
\multirow{6} * {Toys}
 &H@1   &0.0585 &0.0257 &0.0448 &0.0997 &0.1114  &0.1504 &0.1104 &0.1878 &0.1262 &0.1600 &\underline{0.1996} &{0.1935} &\textbf{0.2007}\\
 &H@5   &0.1977 &0.0978 &0.1471 &0.2795 &0.2614  &0.3276 &0.3055 &{0.3682} &0.3344 &0.3389 &0.3613 &\textbf{0.4063} &\underline{0.4033}\\
 &N@5 &0.1286 &0.0614 &0.0960 &0.1919 &0.1885  &0.2423 &0.2102 &0.2820 &0.2327 &0.2528 &{0.2836} &\underline{0.3046} &\textbf{0.3069}\sig\\
 &H@10  &0.3008 &0.1715 &0.2369 &0.3896 &0.3540  &0.4211 &0.4207 &{0.4663} &0.4493 &0.4413 &0.4509 &\underline{0.5062} &\textbf{0.5100}\\
 &N@10&0.1618 &0.0850 &0.1248 &0.2274 &0.2183  &0.2724 &0.2473 &{0.3136} &0.2698 &0.2857 &0.3125 &\underline{0.3368} &\textbf{0.3384}\sig\\
 &MRR    &0.1430 &0.0819 &0.1131 &0.1973 &0.1967  &0.2454 &0.2138 &0.2842 &0.2338 &0.2566 &0.{2871} &\underline{0.3012} &\textbf{0.3048}\sig\\
\midrule
\multirow{6} * {Yelp}
 &H@1   &0.0801 &0.0624 &0.0731 &0.2053 &0.2188  &0.2428 &0.2102 &0.2375 &0.2405 &0.2176 &{0.2493} &\underline{0.2727} &\textbf{0.2816}\msig\\
 &H@5   &0.2415 &0.2036 &0.2249 &0.5437 &0.5111  &0.5768 &0.5707 &0.5745 &{0.5976} &0.5442 &0.5725 &\underline{0.6191} &\textbf{0.6347}\msig\\
 &N@5 &0.1622 &0.1333 &0.1501 &0.3784 &0.3696  &0.4162 &0.3955 &0.4113 &{0.4252} &0.3860 &0.4162 &\underline{0.4527} &\textbf{0.4653}\msig\\
 &H@10  &0.3609 &0.3153 &0.3367 &0.7265 &0.6661  &0.7411 &0.7473 &0.7373 &{0.7597} &0.7096 &0.7371 &\underline{0.7720} &\textbf{0.7863}\msig\\
 &N@10&0.2007 &0.1692 &0.1860 &0.4375 &0.4198  &0.4695 &0.4527 &0.4642 &{0.4778} &0.4395 &0.4696 &\underline{0.5024} &\textbf{0.5146}\msig\\
 &MRR    &0.1740 &0.1470 &0.1616 &0.3630 &0.3595  &0.3988 &0.3751 &0.3927 &{0.4026} &0.3711 &0.4006 &\underline{0.4299} &\textbf{0.4406}\msig\\
\bottomrule
\end{tabular}
}
\end{table*}

The results using the 1-vs-99 test protocol are reported in \autoref{tab:main_table}. To summarize our observations:
\begin{itemize}[leftmargin=*]
    \item Sequential models outperform non-sequential methods such as PopRec, FM and AutoInt, which underscores the importance of item ordering information.
    SAR-based models like SASRec and GCSAN achieve better performance over RNN-based (\eg GRU4Rec), CNN-based (\eg Caser) and GNN-based models (\eg SRGNN), which can be attributed to their lightweight architecture and large receptive field, aligning with criteria (2) and (3) from Section~\ref{sec:crit}. Furthermore, FMLP-Rec outperforms other baseline methods, which could be attributed to the unique sensitivity of its filter layer to item order, substantiating the efficacy of the criterion (1).
    \item ConvFormer significantly outperforms most baseline models across four datasets, with most differences being statistically significant. 
    In addition, the all-convolution architecture of ConvFormer is both computationally efficient and parallelizable, making it efficient for training and inference, as discussed in Appendix~\ref{sec:speed}. Thus, ConvFormer proves to be an effective and efficient solution to sequential user modeling. 
    The superiority of ConvFormer demonstrates that adhering to the criteria proposed, \textit{even a very simple model can outperform many sophisticated solutions and achieve leading performance}, thereby validating the efficacy of the proposed criteria.
\end{itemize}

\subsubsection{Full-sort test on public datasets}\label{sec:full}

Moving forward, We report the full-sort overall performance in \autoref{tab:full}. The results and main observations are consistent with those on the 1-vs-99 test set in \autoref{tab:main_table}. For instance, 
SAR-based models such as SASRec outperform conventional RNN-based models like GRU4Rec and CNN-based models like Caser;
ConvFormer demonstrates the best performance among all the methods. The improvements of ConvFormer are more noticeable on the full-sort test set than on the 1-vs-99 test set due to the full-sort setting being more challenging and providing greater opportunities for improvement. 
\begin{table}[t!]
	\caption{Full-sort performance of different methods on four datasets. The best and second performance methods are marked in bold and underlined fonts, respectively.}	\label{tab:full}
    \centering
    \small
	\setlength{\tabcolsep}{5.5mm}{
	\begin{tabular}{ccccccc}
	\toprule
		Datasets &Metric &GRU4Rec &Caser &SASRec &FMLP-Rec &ConvFormer\\
	\midrule
\multirow{6} * {Beauty}
&H@5 & 0.0164 & 0.0205 & 0.0387 & \underline{0.0398} & \textbf{0.0413}\\
&N@5 & 0.0099 & 0.0131 & 0.0249 & \underline{0.0258} & \textbf{0.0270}\\
&H@10& 0.0283 & 0.0347 & 0.0605 & \underline{0.0632} & \textbf{0.0675}\\
&N@10& 0.0137 & 0.0176 & 0.0318 & \underline{0.0333} & \textbf{0.0354}\\
&H@20& 0.0479 & 0.0556 & 0.0902 & \underline{0.0958} & \textbf{0.0993}\\
&N@20& 0.0187 & 0.0229 & 0.0394 & \underline{0.0415} & \textbf{0.0433}\\
\midrule
\multirow{6} * {Sports}
&H@5 & 0.0129 & 0.0116 & \underline{0.0233} & 0.0218 & \textbf{0.0244}\\
&N@5 & 0.0086 & 0.0072 & \underline{0.0154} & 0.0144 & \textbf{0.0157}\\
&H@10& 0.0204 & 0.0194 & \underline{0.0350} & 0.0344 & \textbf{0.0387}\\
&N@10& 0.0110 & 0.0097 & \underline{0.0192} & 0.0185 & \textbf{0.0203}\\
&H@20& 0.0333 & 0.0314 & 0.0507 & \underline{0.0537} & \textbf{0.0587}\\
&N@20& 0.0142 & 0.0126 & 0.0231 & \underline{0.0233} & \textbf{0.0253}\\
\midrule
\multirow{6} * {Toys}
&H@5 & 0.0097 & 0.0166 & \underline{0.0463} & 0.0456 & \textbf{0.0502}\\
&N@5 & 0.0059 & 0.0107 & 0.0306 & \underline{0.0317} & \textbf{0.0344}\\
&H@10& 0.0176 & 0.0270 & 0.0675 & \underline{0.0683} & \textbf{0.0753}\\
&N@10& 0.0084 & 0.0141 & 0.0374 & \underline{0.0391} & \textbf{0.0424}\\
&H@20& 0.0301 & 0.0420 & 0.0941 & \underline{0.0991} & \textbf{0.1056}\\
&N@20& 0.0116 & 0.0179 & 0.0441 & \underline{0.0468} & \textbf{0.0500}\\
\midrule
\multirow{6} * {Yelp}
&H@5 & 0.0152 & 0.0151 & 0.0162 & \underline{0.0179} & \textbf{0.0212}\\
&N@5 & 0.0099 & 0.0096 & 0.0100 & \underline{0.0113} & \textbf{0.0137}\\
&H@10& 0.0263 & 0.0253 & 0.0274 & \underline{0.0304} & \textbf{0.0353}\\
&N@10& 0.0134 & 0.0129 & 0.0136 & \underline{0.0153} & \textbf{0.0182}\\
&H@20& 0.0439 & 0.0422 & 0.0457 & \underline{0.0511} & \textbf{0.0566}\\
&N@20& 0.0178 & 0.0171 & 0.0182 & \underline{0.0205} & \textbf{0.0235}\\

\bottomrule
	\end{tabular}
	}
\end{table}

\subsubsection{Full-sort test on large-scale industrial dataset}\label{sec:industrial}
\begin{table}
\centering
\caption{Performance comparison on our Xbox dataset. Bold and underlined fonts indicate the first and second best results, respectively.}\label{tab:industry}
    \setlength{\tabcolsep}{5mm}{
    \small
    \begin{tabular}{lllllll}
    \toprule
    Model  &HIT@10 & HIT@20 & HIT@30 & NDCG@10 & NDCG@20 & NDCG@30 \\ \midrule
    GRU4Rec& 0.5732 & 0.6797 & 0.7365 & 0.3867 & 0.4137 & 0.4258 \\
    SASRec & 0.5635 & 0.6722 & 0.7317 & 0.3817 & 0.4092 & 0.4218 \\
    FMLP-Rec&\underline{0.5781} & \underline{0.6861} & \underline{0.7449} & \underline{0.3903} & \underline{0.4177} & \underline{0.4302} \\
    ConvFormer&\textbf{0.5996} & \textbf{0.7078} & \textbf{0.7645} & \textbf{0.4002} & \textbf{0.4276} & \textbf{0.4397} \\\bottomrule
\end{tabular}}
\end{table}
\begin{table}[t!]
	\caption{Full-sort performance on the Xbox dataset. The best results are marked in bold fonts. Increasing the dropout rate consistently leads to performance drop.}	\label{tab:full2}
    \centering
    \small
	\setlength{\tabcolsep}{6.5mm}{
	\begin{tabular}{cccccc}
	\toprule
		Methods & Metric & dropout=0.0 & dropout=0.1 & dropout=0.3 & dropout=0.5 \\
	\midrule
\multirow{6} * {SASRec}
&H@5 & 0.5635 & \textbf{0.5662} & 0.5256 & 0.4976\\
&N@5 & 0.6722 & \textbf{0.6747} & 0.6339 & 0.6092\\
&H@10& 0.7317 & \textbf{0.7335} & 0.6948 & 0.6723\\
&N@10& \textbf{0.3817} & 0.3751 & 0.3483 & 0.3222\\
&H@20& \textbf{0.4092} & 0.4026 & 0.3757 & 0.3504\\
&N@20& \textbf{0.4218} & 0.4152 & 0.3887 & 0.3638\\
\midrule
\multirow{6} * {ConvFormer}
&H@5 & \textbf{0.5996} & 0.5836 & 0.5661 & 0.5513\\
&N@5 & \textbf{0.7078} & 0.6937 & 0.6796 & 0.6650\\
&H@10& \textbf{0.7645} & 0.7519 & 0.7395 & 0.7267\\
&N@10& \textbf{0.4002} & 0.3879 & 0.3720 & 0.3569\\
&H@20& \textbf{0.4276} & 0.4158 & 0.4008 & 0.3857\\
&N@20& \textbf{0.4397} & 0.4282 & 0.4136 & 0.3989\\
\bottomrule
	\end{tabular}
	}
\end{table}
Since the public datasets above have limited scale, there is a possibility that the superiority of ConvFormer is due to Transformer's overfitting on these datasets rather than the inefficacy of Transformer's item-to-item paradigm for sequential user modeling. 
To address this concern, it is crucial to evaluate ConvFormer against competitive baselines, especially Transformer (SASRec), using large-scale industrial datasets.

To this end, we construct an industrial dataset using the real-world gaming recommendation log on the Xbox platform. 
The results on the Xbox dataset are present in \autoref{tab:industry} and \autoref{tab:full2}. Notably, we observed that GRU4Rec outperforms SASRec, which suggests that the evolving of user preference is significant in real-world scenarios, thus highlighting the importance of our criterion (1) regarding order-sensitivity.
Moreover, ConvFormer achieved a significantly better performance compared to other baselines in particular SASRec. These findings confirm the superiority of ConvFormer and the efficacy of the three criteria we proposed, in large-scale real-world applications.

\paragraph{Safeguards.}{The desensitized and encrypted dataset contains no Personal Identifiable Information (PII). Adequate data protection was carried out during experiment to prevent the risk of data copy leakage. The dataset does not represent any business situation, only used for academic research.}

\subsection{Ablation studies}\label{sec:ablation}
    \begin{figure}
    \centering
    \includegraphics[width=0.49\linewidth, trim=15 10 10 10]{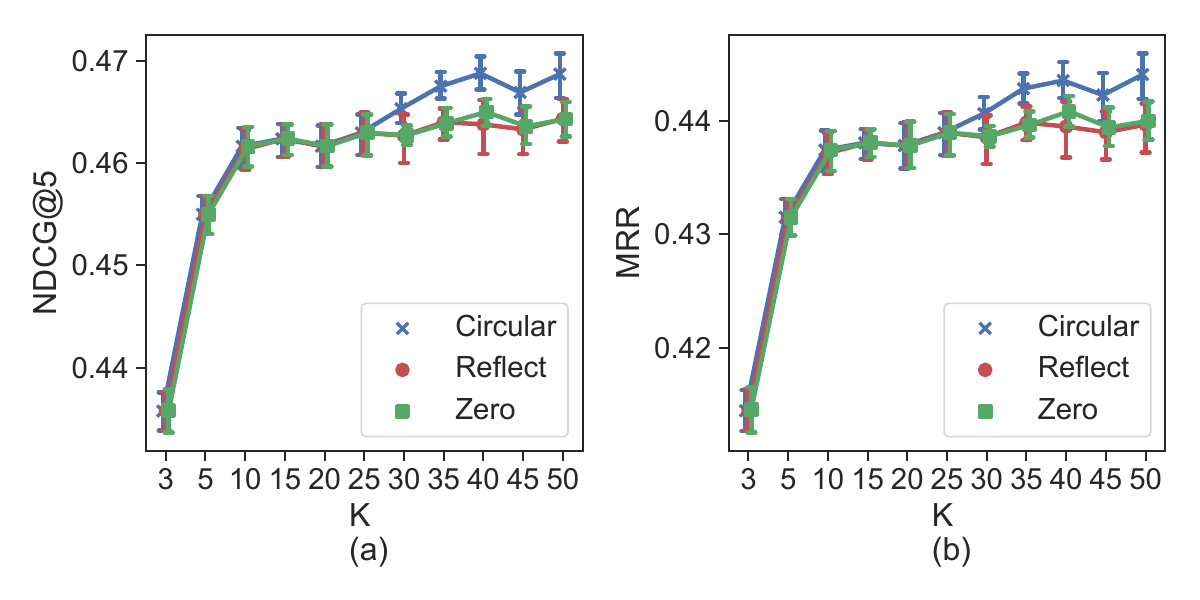}
    \includegraphics[width=0.49\linewidth, trim=10 10 15 10]{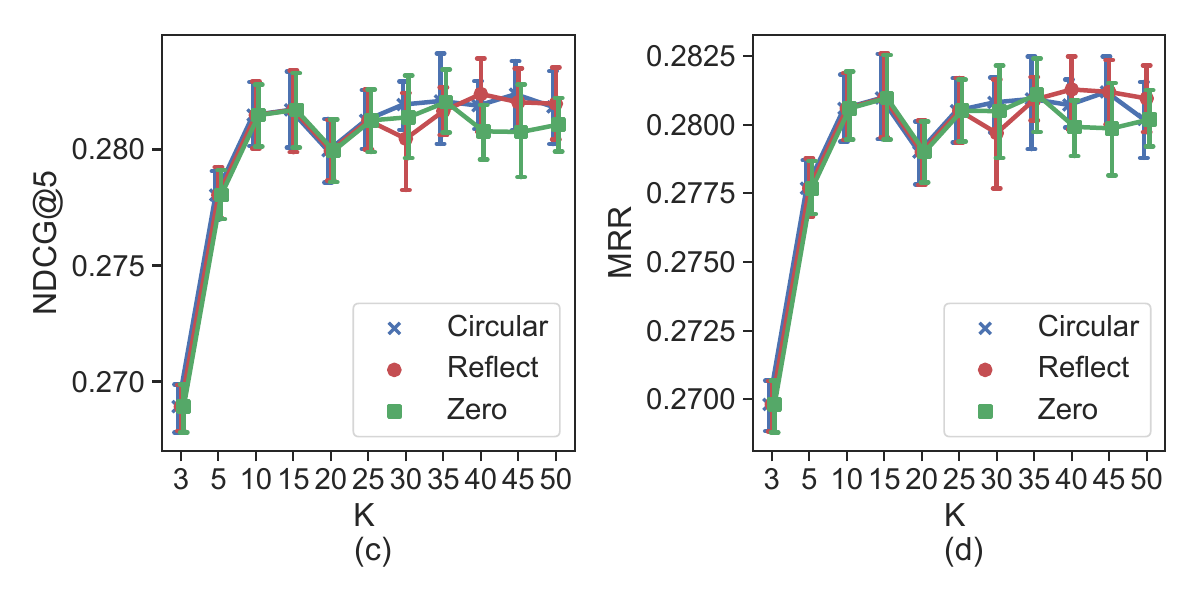}
    \caption{Impact of Convformer's receptive field size K on model performance over Yelp (a-b) and Sports (c-d) datasets. Error bar denotes 95\% confidence interval.}\label{fig:param_k}
\end{figure}

We have showcased the effectiveness of the three criteria through the superior performance of ConvFormer.  
In this section, we deconstruct it to further assess the role of each criterion. 
\subsubsection{Large receptive field}
To demonstrate the necessity of a large receptive field, we change the kernel size $\mathrm{K}$ to visualize its impact in \autoref{fig:param_k}. Results show that an increase in kernel size leads to improved performance, as evidenced by the rise in MRR from 0.414 at $\mathrm{K}=3$ to approximately 0.441 at $\mathrm{K}=50$ on Yelp.

We also investigate the role of padding in the convolution operator, which are denoted by \textsl{Circular}, \textsl{Reflect}, and \textsl{Zero} in \autoref{fig:param_k}. Specifically, in scenarios with strong behavior periodicity (such as Yelp), circular padding, which preserves the periodic property, performs significantly better than other padding methods. However, in scenarios with weak periodicity in user behaviors (such as Amazon Sports), the performance difference between padding methods is minimal.

\subsubsection{Lightweight convolution}\label{sec:lightconv}

To verify the role of lightweight architecture, we replace the LighTCN operator (denoted by Conv-L) of ConvFormer with two variants: the vanilla convolution operator (denoted by Conv-V) and the separable convolution operator~\cite{DBLP:journals/corr/mobilenet} (denoted by Conv-S).
Notably, both Conv-V and Conv-S meet criteria (1) and (2), but fail criterion (3).

\begin{wrapfigure}{r}{6.5cm}
    \vspace{-0.2cm}
    \centering
    \includegraphics[width=\linewidth, trim=30 20 30 20]{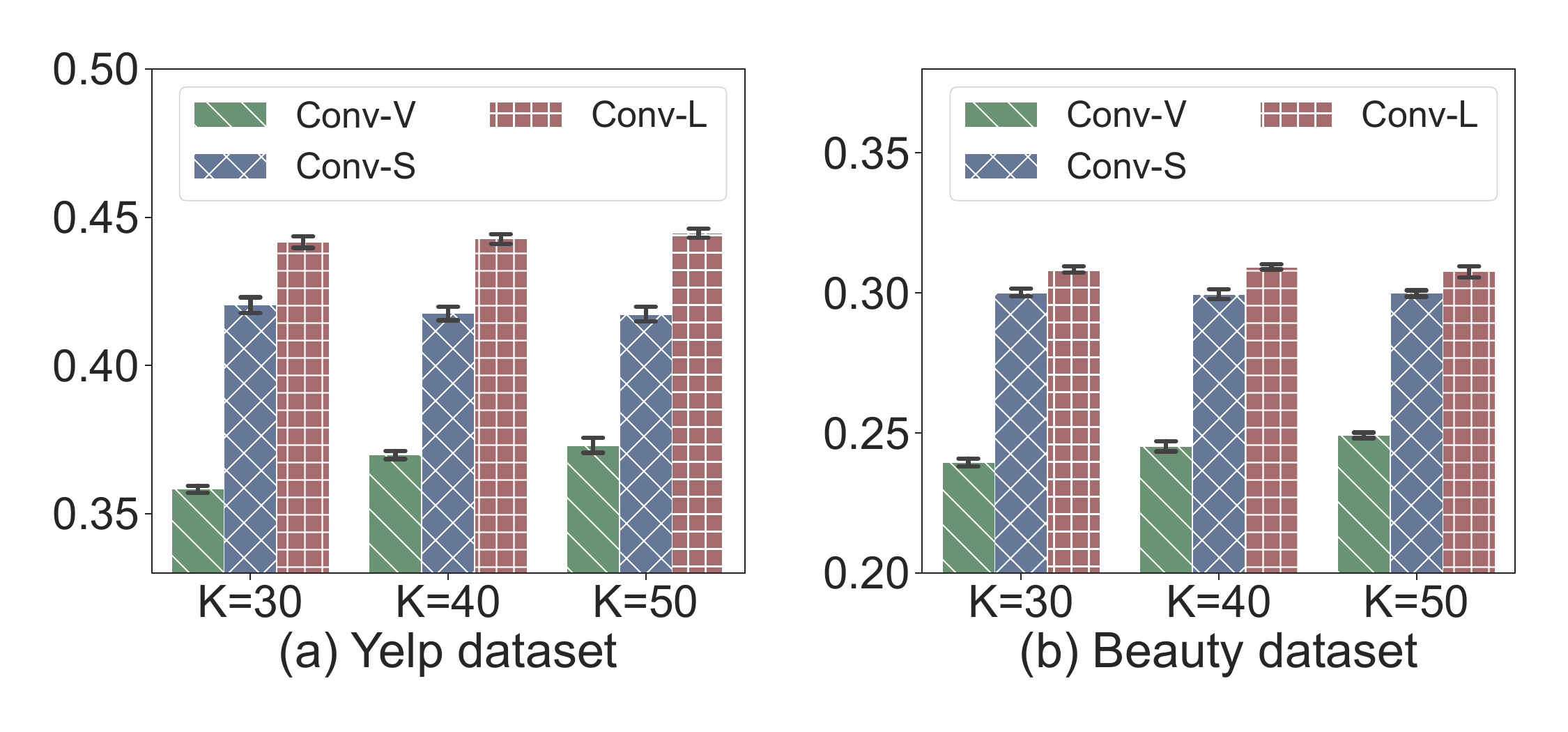}
    \caption{Impact of lightweight convolution.}\label{fig:paramshared_conv}
    \vspace{-0.2cm}
\end{wrapfigure}
According to \autoref{fig:paramshared_conv}, the Conv-L operator (used in our standard ConvFormer) largely outperforms the vanilla convolution operator due to its suppression of over-parameterization. Specifically, when K=30, it improves the MRR by a relative 23.24\% on Yelp and 28.71\% on Beauty.
The Conv-S operator also performs better than the vanilla convolution operator, but worse than Conv-L. Its inferiority is attributed to the extra interchannel interaction compared to Conv-L. This redundancy increases the risk of overparameterization, as the subsequent FFN modules are specifically designed for interchannel interactions.

\subsubsection{Attention vs. convolution}\label{sec:attnconv}

To support the claim that self-attentive modules can be a hindrance for sequence recommendation due to the insensitivity of item order, we replace the LighTCN module with variants of attentive mechanisms. We select to compare ConvFormer with Fastformer \cite{wu2021fastformer} and PoolingFormer \cite{zhang2021poolingformer}, as these two models beat a series of efficient Transformer variants such as LinFormer~\cite{wang2020linformer} and LongFormer~\cite{beltagy2020longformer}.  Notably, both additional baselines satisfy criteria (2) and (3), i.e., having large receptive fields and lightweight architectures, but fail to meet criterion (1), i.e., they are developed based on attentive paradigms that is insensitive to item order.

\begin{table}
\caption{Comparison with emerging attentive light-weight methods. 
The bold fonts represent the best performance. “*” marks the metrics that ConvFormer improves significantly over the best baselines, with p-value $<$ 0.01 in the paired sample t-test.
} 
\label{tab:itemtoitem_conv}
\setlength{\tabcolsep}{5mm}{
\small
\begin{tabular}{llcccccc}
	\toprule
	Dataset                 & Model & H@1             & H@5             & H@10             & N@5              & N@10             & MRR              \\ 
        \midrule
\multirow{5}{*}{Beauty} & SASRec  & 0.1870 & 0.3741 & 0.4696 & 0.2848 & 0.3156 & 0.2852           \\
                        & FastFormer & 0.1405 & 0.3395 & 0.4454 & 0.2438 & 0.2780 & 0.2449  \\
                        & PoolingFormer & 0.1930 & 0.3932  & 0.4925 & 0.2981 & 0.3302 & 0.2971           \\
                        & ConvFormer & \textbf{0.2019}\sig & \textbf{0.4119}\sig & \textbf{0.5105}\sig & \textbf{0.3125}\sig & \textbf{0.3443}\sig & \textbf{0.3093}\sig      \\                   
                        \midrule
\multirow{5}{*}{Sports} & SASRec   & 0.1445 & 0.3466 & 0.4622 & 0.2497 & 0.2869 & 0.2520           \\
                        & FastFormer & 0.1185 & 0.3249 & 0.4573 & 0.2238 & 0.2665 & 0.2284 \\
                        & PoolingFormer & 0.1568 & 0.3741 & 0.5000 & 0.2687 & 0.3093 & 0.2693           \\
                        & ConvFormer & \textbf{0.1671}\sig & \textbf{0.3891}\sig & \textbf{0.5116}\sig & \textbf{0.2819}\sig & \textbf{0.3215}\sig & \textbf{0.2808}\sig           \\   
                       \midrule
\multirow{5}{*}{Toys} & SASRec   & 0.1878 & 0.3682 & 0.4663 & 0.2820 & 0.3136 & 0.2842           \\
                        & FastFormer & 0.1301 & 0.3390 & 0.4517 & 0.2380 & 0.2744 & 0.2384 \\
                        & PoolingFormer & 0.1893 & 0.3873 & 0.4893 & 0.2927 & 0.3256 & 0.2925           \\
                        & ConvFormer & \textbf{0.2007}\sig & \textbf{0.4033}\sig & \textbf{0.5100}\sig & \textbf{0.3069}\sig & \textbf{0.3384}\sig & \textbf{0.3048}\sig           \\           
                       \midrule
\multirow{5}{*}{Yelp} & SASRec   & 0.2375 & 0.5745 & 0.7373 & 0.4113 & 0.4642 & 0.3927           \\
                        & FastFormer & 0.1918 & 0.5451 & 0.7355 & 0.3727 & 0.4344 & 0.3557 \\
                        & PoolingFormer & 0.2539 & 0.6087 & 0.7663 & 0.4378 & 0.4890 & 0.4144           \\
                        & ConvFormer & \textbf{0.2816}\sig & \textbf{0.6347}\sig & \textbf{0.7863}\sig & \textbf{0.4653}\sig & \textbf{0.5146}\sig & \textbf{0.4406}\sig           \\                  
\bottomrule
\end{tabular}
}
\end{table}
According to \autoref{tab:itemtoitem_conv}, emerging Transformer variants with large receptive fields and more lightweight architectures could achieve performance gains in sequential user modeling. For example, PoolingFormer improves MRR by approximately 4.2\% over SASRec. However, the gap between these methods and ConvFormer remains significant. As a result, it is reasonable to conclude that the item-to-item paradigm in Transformer is a bottleneck for user behavior understanding, potentially due to the lack of order sensitivity. 

\subsection{Verification of acceleration and approximation}\label{sec:speed}
\begin{figure}[!ht]
    \begin{minipage}{0.49\linewidth}
    \includegraphics[width=\linewidth, trim=30 20 30 20]{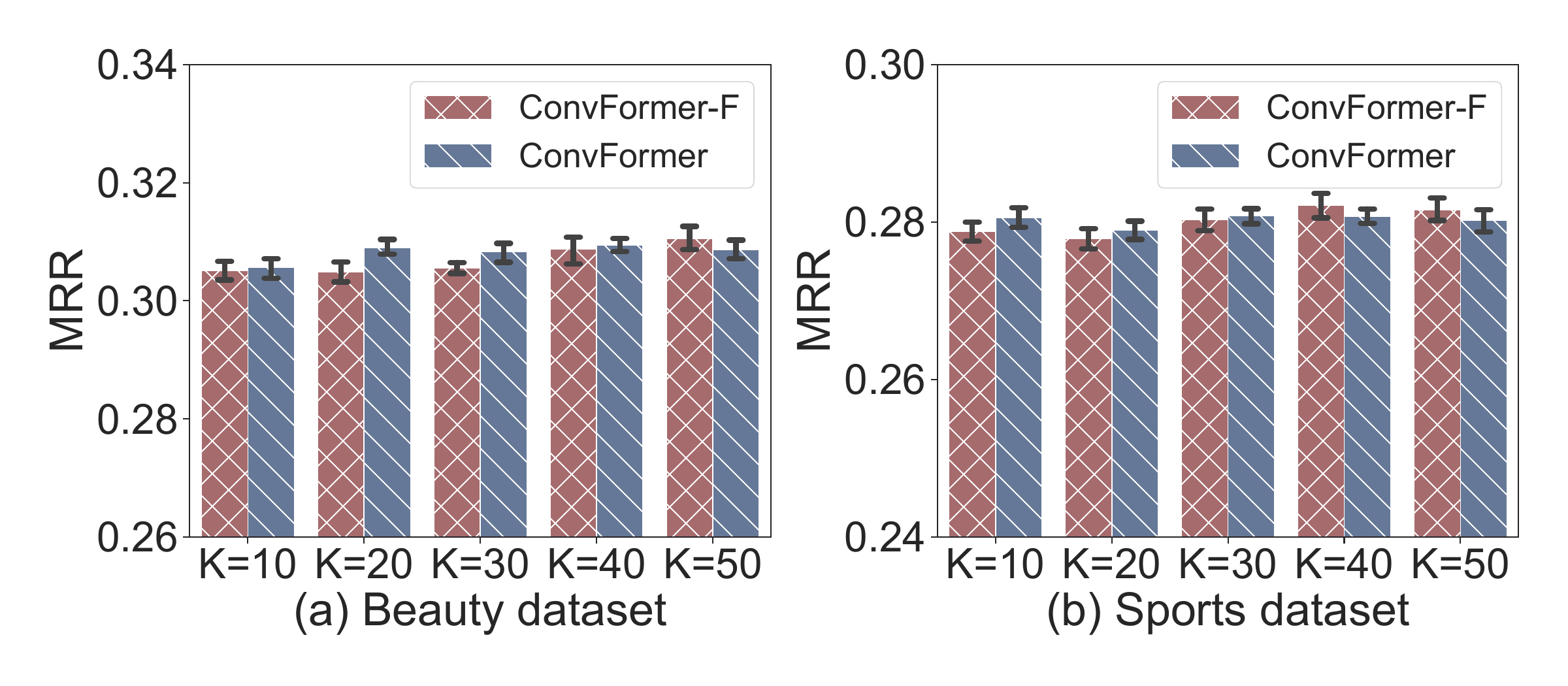}
    \caption{Comparing the accuracy achieved by ConvFormer and ConvFormer-F at different kernel size $\mathrm{K}$.}\label{fig:acc}
    \end{minipage}
    \hfill
    \begin{minipage}{0.49\linewidth}
    \includegraphics[width=\linewidth, trim=15 20 15 5]{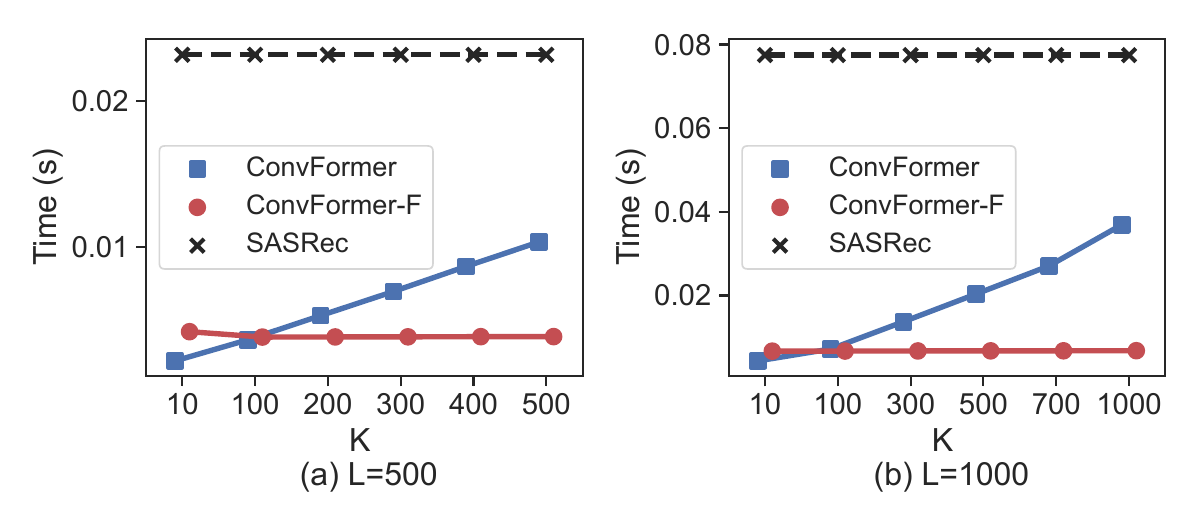}
    \caption{Comparing the inference time with different sequence length $\mathrm{L}$ and kernel size $\mathrm{K}$.}\label{fig:speed}
    \end{minipage}
\end{figure}

To showcase the efficacy of ConvFormer-F, it is necessary to verify its accuracy equivalence and speed acceleration with respect to the standard ConvFormer.
Overall, we reuse the hyperparameters in \autoref{tab:param}, but for the stability of test results, we set the batch size to 512. 
To emphasize the difference in speed, we omit the inference time of these methods' common layers, including embedding and FFN layers.
In practice, we firstly generate a random matrix $\bR\in\mathbb{R}^{\mathrm{L}\times\mathrm{D}}$ and then feed it into a self-attention layer, a CWC layer and its accelerated version.
We vary the maximum sequence length $\mathrm{L}$ in \{500, 1000\} and the convolution kernel size from 10 to $\mathrm{L}$, and record the average GPU inference time over 10 runs.
Experiments are conducted with an AMD EPYC 7742 64-Core processor and an NVIDIA RTX A6000 GPU.

The accuracy comparison is conducted in \autoref{fig:acc}. Overall, there is no significant difference between the two methods, as evidenced by the substantial overlap between the 95\% confidence intervals represented by the error bars.
Precisely, the MRR differences at $K=40$ are merely $8e^{-4}$ and $1e^{-3}$ for the beauty and sports datasets, respectively.
These observations support the accuracy equivalence between ConvFormer and ConvFormer-F.

The GPU inference time is compared in \autoref{fig:speed}. 
The CPU inference time follows similar patterns, therefore it has been omitted. 
The y-axis indicates the total inference time for a batch of 512 sequences. Overall, the inference time of ConvFormer rises linearly with respect to the kernel size $\mathrm{K}$, while that of ConvFormer-F keeps constant with respect to K.  As a result, the speedup of the fast approximation approach is not readily apparent for small kernel sizes, \eg $\mathrm{K}<100$ in $\mathrm{L}=1000$; nonetheless, as the kernel size is increased for better accuracy, the superiority of ConvFormer-F becomes more pronounced.
Note that the inference cost is a major flaw with SASRec, which is mostly brought on by its item-to-item paradigm and softmax operator. In particular, ConvFormer and ConvFormer-F accelerate SASRec by 3x and 5x, respectively, even with the largest kernel setting, \ie $\mathrm{K}=\mathrm{L}$.

\section{Related Works}
\paragraph{Sequential recommendation.}
Recommendation system aims to capture the key preference of users from their profiles and historical interactions~\cite{uplift1,xdeepfm,mf1,mf2}. In parallel to other recommendation tasks such as collaborative filtering~\cite{he2017neural},  cold-starting~\cite{braunhofer2014switching}, unbiased learning~\cite{escm2,debias1,debias2,debias3,escm2j}, and pretraining~\cite{hou2022towards}, sequential recommendation, which is typically formulated as a next-item-prediction problem, is also one of the foundation tasks for RS~\cite{caser,sasrec,bert4rec,zhou2019deep,adaranker,fmlp}. The challenges lie in how to precisely mine evolving preference patterns from users' behavior sequence (aka sequential user modeling) and then predict what users will be interested in soon.
The fundamental aspect of sequential user modeling is the handling of sequences, therefore neural architectures that are able to model sequences, e.g., RNN~\cite{DBLP:journals/corr/HidasiKBT15} and CNN~\cite{caser,cnn,yuan2020parameter}, can be utilized in this area. 
Similar architectures include \cite{zhou2019deep,lian2021multi,wu2017recurrent,ren2019lifelong,memory1}, with extended modules such as target-aware attention, memory networks and hierarchical RNNs. 
CNNs have also been used for sequential user modelings, as a sequence of items can be treated as a $1$-D image and local features can be encoded with convolution filters.  
Motivated by this, \cite{caser,cnn,yuan2020parameter} propose to use CNN for sequential user modelings. 
Some researchers argue that an individual item sequence cannot fully reveal a user's behavior patterns, so they construct a graph of items which contains more information than sequence, and apply graph neural networks for sequential user modeling~\cite{srgnn, gcsan,10.1145/3357384.3358010,yuan2020parameter}. 
Attention mechanism, despite its simplicity, has proven to be effective for sequential user modeling~\cite{ying2018sequential,zhang2022efficiently,niu2021review}. The Transformer~\cite{vaswani2017attention}, as a generic style for self-attention mechanism, has been another foundational model in sequential user modeling. 
Typical works include \cite{sasrec,sasrec2,bert4rec,chen2019behavior}. 
However, most of the works in this line simply apply the Transformer architecture to the field of recommendation, without further identifying the key difference between this field with other fields such as NLP. 
Recently, researchers~\cite{mlp4rec,fmlp} show that the vanilla Transformer architecture may not be optimal for effective sequential user modeling, and alternative structures such as all-MLP architectures have demonstrated promising performance. Motivate by these factors, this paper aims to further investigate the essential criteria for sequential user modeling.

\paragraph{Transformer applications and alternatives.}
Starting with BERT~\cite{DBLP:conf/naacl/DevlinCLT19}, the Transformer architecture and pretraining paradigm have gained widespread popularity in NLP due to their remarkable performance and are now spreading to other domains, e.g., Swin Transformer in CV~\cite{liu2021swin}, BEiT-v3 in vision-language tasks~\cite{wang2022image}, and AlphaFold-v2 in AI4Science~\cite{jumper2021highly}. It is interesting to note that the success of these applications follows a similar trend: starting with a direct application and then adapting to specific domains. We believe that it is time to rethink the unique adaptation for recommender systems. 
On the other hand, several prior works have investigated the feasibility of the alternatives of Transformer. 
For example, researchers found that replacing the self-attention matrix with a parameter matrix could improve overall performance~\cite{tay2005synthesizer} . They also concluded that a random initialized matrix is a competitive alternative. Similarly, MLP-Mixer~\cite{mlpmixer}, which replaces self-attention with fixed learnable weights, has shown advantages over canonical Transformer in many fundamental and data-rich fields. 
These findings inspire to develop alternatives to Transformer in specific domains, leveraging domain data and task characteristics, to achieve better overall performance.

\section{Conclusions}
In this study, we re-evaluate Transformer-like architectures for sequential user modeling and identify three critical criteria for effective token mixers. Guided by these criteria, we develop ConvFormer, a streamlined modification of the Transformer, augmented with an acceleration technique for computational efficiency. Our findings demonstrate that even a simplified model, when designed in accordance with these criteria, can outperform various complex and delicate solutions, thereby validating the efficacy of the proposed criteria.

\textbf{Limitations.} 
We construct the proposed criteria using the standard two-tower architecture for item retrieval in sequential user modeling.While ConvFormer's effectiveness is also confirmed in a one-tower setup (detailed in the appendix), a thorough examination of each criterion in this context is pending and warrants future study. Notably, ConvFormer primarily serves as a proof-of-concept, leaving ample scope for future research to explore a broader range of sequential models and tasks to validate the proposed criteria.

\bibliography{bib}

\begin{thebibliography}{10}

\bibitem{beltagy2020longformer}
Iz~Beltagy, Matthew~E Peters, and Arman Cohan.
\newblock Longformer: The long-document transformer.
\newblock {\em arXiv preprint arXiv:2004.05150}, 2020.

\bibitem{braunhofer2014switching}
Matthias Braunhofer, Victor Codina, and Francesco Ricci.
\newblock Switching hybrid for cold-starting context-aware recommender systems.
\newblock In {\em RecSys}, pages 349--352, 2014.

\bibitem{chen2021end}
Qiwei Chen, Changhua Pei, Shanshan Lv, Chao Li, Junfeng Ge, and Wenwu Ou.
\newblock End-to-end user behavior retrieval in click-through rateprediction
  model.
\newblock {\em arXiv preprint arXiv:2108.04468}, 2021.

\bibitem{chen2019behavior}
Qiwei Chen, Huan Zhao, Wei Li, Pipei Huang, and Wenwu Ou.
\newblock Behavior sequence transformer for e-commerce recommendation in
  alibaba.
\newblock In {\em Proceedings of the 1st International Workshop on Deep
  Learning Practice for High-Dimensional Sparse Data}, pages 1--4, 2019.

\bibitem{DBLP:conf/naacl/DevlinCLT19}
Jacob Devlin, Ming{-}Wei Chang, Kenton Lee, and Kristina Toutanova.
\newblock {BERT:} pre-training of deep bidirectional transformers for language
  understanding.
\newblock In Jill Burstein, Christy Doran, and Thamar Solorio, editors, {\em
  NAACL}, pages 4171--4186, 2019.

\bibitem{adaranker}
Xinyan Fan, Jianxun Lian, Wayne~Xin Zhao, Zheng Liu, Chaozhuo Li, and Xing Xie.
\newblock Ada-ranker: {A} data distribution adaptive ranking paradigm for
  sequential recommendation.
\newblock In {\em SIGIR}, pages 1599--1610, 2022.

\bibitem{10.1145/2843948}
Carlos~A. Gomez-Uribe and Neil Hunt.
\newblock The netflix recommender system: Algorithms, business value, and
  innovation.
\newblock {\em ACM Trans. Manage. Inf. Syst.}, 6(4), dec 2016.

\bibitem{he2017neural}
Xiangnan He, Lizi Liao, Hanwang Zhang, Liqiang Nie, Xia Hu, and Tat-Seng Chua.
\newblock Neural collaborative filtering.
\newblock In {\em WWW}, pages 173--182, 2017.

\bibitem{gru}
Balazs Hidasi and Alexandros Karatzoglou.
\newblock Recurrent neural networks with top-k gains for session-based
  recommendations.
\newblock In {\em CIKM}, pages 843--852, 2018.

\bibitem{DBLP:journals/corr/HidasiKBT15}
Bal{\'{a}}zs Hidasi, Alexandros Karatzoglou, Linas Baltrunas, and Domonkos
  Tikk.
\newblock Session-based recommendations with recurrent neural networks.
\newblock In Yoshua Bengio and Yann LeCun, editors, {\em ICLR}, 2016.

\bibitem{hou2022towards}
Yupeng Hou, Shanlei Mu, Wayne~Xin Zhao, Yaliang Li, Bolin Ding, and Ji-Rong
  Wen.
\newblock Towards universal sequence representation learning for recommender
  systems.
\newblock In {\em SIGKDD}, pages 585--593, 2022.

\bibitem{DBLP:journals/corr/mobilenet}
Andrew~G. Howard, Menglong Zhu, Bo~Chen, Dmitry Kalenichenko, Weijun Wang,
  Tobias Weyand, Marco Andreetto, and Hartwig Adam.
\newblock Mobilenets: Efficient convolutional neural networks for mobile vision
  applications.
\newblock {\em CoRR}, abs/1704.04861, 2017.

\bibitem{memory1}
Jin Huang, Wayne~Xin Zhao, Hongjian Dou, Ji{-}Rong Wen, and Edward~Y. Chang.
\newblock Improving sequential recommendation with knowledge-enhanced memory
  networks.
\newblock In Kevyn Collins{-}Thompson, Qiaozhu Mei, Brian~D. Davison, Yiqun
  Liu, and Emine Yilmaz, editors, {\em SIGIR}, pages 505--514. {ACM}, 2018.

\bibitem{hgn}
Kexin Huang, Ye~Du, Li~Li, Jun Shen, and Geng Sun.
\newblock Pairwise-based hierarchical gating networks for sequential
  recommendation.
\newblock In {\em KSEM}, volume 12275, pages 64--75, 2020.

\bibitem{jumper2021highly}
John Jumper, Richard Evans, Alexander Pritzel, Tim Green, Michael Figurnov,
  Olaf Ronneberger, Kathryn Tunyasuvunakool, Russ Bates, et~al.
\newblock Highly accurate protein structure prediction with alphafold.
\newblock {\em Nature}, 596(7873):583--589, 2021.

\bibitem{sasrec}
Wang{-}Cheng Kang and Julian~J. McAuley.
\newblock Self-attentive sequential recommendation.
\newblock In {\em ICDM}, pages 197--206, 2018.

\bibitem{lee2022fnet}
James Lee-Thorp, Joshua Ainslie, Ilya Eckstein, and Santiago Ontanon.
\newblock Fnet: Mixing tokens with fourier transforms.
\newblock In {\em NAACL}, pages 4296--4313, 2022.

\bibitem{debias1}
Haoxuan Li, Yanghao Xiao, Chunyuan Zheng, and Peng Wu.
\newblock Balancing unobserved confounding with a few unbiased ratings in
  debiased recommendations.
\newblock In {\em {WWW}}, pages 1305--1313. {ACM}, 2023.

\bibitem{debias2}
Haoxuan Li, Chunyuan Zheng, and Peng Wu.
\newblock Stabledr: Stabilized doubly robust learning for recommendation on
  data missing not at random.
\newblock In {\em {ICLR}}. OpenReview.net, 2023.

\bibitem{sasrec2}
Jiacheng Li, Yujie Wang, and Julian~J. McAuley.
\newblock Time interval aware self-attention for sequential recommendation.
\newblock In James Caverlee, Xia~(Ben) Hu, Mounia Lalmas, and Wei Wang,
  editors, {\em WSDM}, pages 322--330, 2020.

\bibitem{mlp4rec}
Muyang Li, Xiangyu Zhao, Chuan Lyu, Minghao Zhao, Runze Wu, and Ruocheng Guo.
\newblock Mlp4rec: {A} pure {MLP} architecture for sequential recommendations.
\newblock In {\em {IJCAI}}, pages 2138--2144. ijcai.org, 2022.

\bibitem{lian2021multi}
Jianxun Lian, Iyad Batal, Zheng Liu, Akshay Soni, Eun~Yong Kang, Yajun Wang,
  and Xing Xie.
\newblock Multi-interest-aware user modeling for large-scale sequential
  recommendations.
\newblock {\em arXiv preprint arXiv:2102.09211}, 2021.

\bibitem{xdeepfm}
Jianxun Lian, Xiaohuan Zhou, Fuzheng Zhang, Zhongxia Chen, Xing Xie, and
  Guangzhong Sun.
\newblock xdeepfm: Combining explicit and implicit feature interactions for
  recommender systems.
\newblock In {\em KDD}, pages 1754--1763, 2018.

\bibitem{mf2}
Zhipeng Lin, Wenjing Yang, Yongjun Zhang, Haotian Wang, and Yuhua Tang.
\newblock Mulattenrec: {A} multi-level attention-based model for
  recommendation.
\newblock In {\em {ICONIP} {(2)}}, volume 11302 of {\em Lecture Notes in
  Computer Science}, pages 240--252. Springer, 2018.

\bibitem{liu2020kalman}
Hu~Liu, Jing Lu, Xiwei Zhao, Sulong Xu, Hao Peng, Yutong Liu, Zehua Zhang, Jian
  Li, Junsheng Jin, Yongjun Bao, et~al.
\newblock Kalman filtering attention for user behavior modeling in ctr
  prediction.
\newblock {\em NeurIPS}, 33:9228--9238, 2020.

\bibitem{liu2021swin}
Ze~Liu, Yutong Lin, Yue Cao, Han Hu, Yixuan Wei, Zheng Zhang, Stephen Lin, and
  Baining Guo.
\newblock Swin transformer: Hierarchical vision transformer using shifted
  windows.
\newblock In {\em ICCV}, pages 10012--10022, 2021.

\bibitem{DBLP:conf/sigir/McAuleyTSH15}
Julian~J. McAuley, Christopher Targett, Qinfeng Shi, and Anton van~den Hengel.
\newblock Image-based recommendations on styles and substitutes.
\newblock In {\em SIGIR}, pages 43--52, 2015.

\bibitem{niu2021review}
Zhaoyang Niu, Guoqiang Zhong, and Hui Yu.
\newblock A review on the attention mechanism of deep learning.
\newblock {\em Neurocomputing}, 452:48--62, 2021.

\bibitem{dsp}
Alan~V Oppenheim, John~R Buck, and Ronald~W Schafer.
\newblock {\em Discrete-time signal processing. Vol. 2}.
\newblock Upper Saddle River, NJ: Prentice Hall, 2001.

\bibitem{pi2020search}
Qi~Pi, Guorui Zhou, Yujing Zhang, Zhe Wang, Lejian Ren, Ying Fan, Xiaoqiang
  Zhu, and Kun Gai.
\newblock Search-based user interest modeling with lifelong sequential behavior
  data for click-through rate prediction.
\newblock In {\em CIKM}, pages 2685--2692, 2020.

\bibitem{fcnn}
Harry Pratt, Bryan Williams, Frans Coenen, and Yalin Zheng.
\newblock Fcnn: Fourier convolutional neural networks.
\newblock In {\em PKDD}, pages 786--798. Springer, 2017.

\bibitem{clea}
Yuqi Qin, Pengfei Wang, and Chenliang Li.
\newblock The world is binary: Contrastive learning for denoising next basket
  recommendation.
\newblock In {\em SIGIR}, pages 859--868, 2021.

\bibitem{10.1145/3357384.3358010}
Ruihong Qiu, Jingjing Li, Zi~Huang, and Hongzhi YIn.
\newblock Rethinking the item order in session-based recommendation with graph
  neural networks.
\newblock In {\em CIKM}, page 579–588, 2019.

\bibitem{gfnet}
Yongming Rao, Wenliang Zhao, Zheng Zhu, Jiwen Lu, and Jie Zhou.
\newblock Global filter networks for image classification.
\newblock In {\em NeurIPS}, volume~34, pages 980--993, 2021.

\bibitem{ren2019lifelong}
Kan Ren, Jiarui Qin, Yuchen Fang, Weinan Zhang, Lei Zheng, Weijie Bian, Guorui
  Zhou, Jian Xu, Yong Yu, Xiaoqiang Zhu, et~al.
\newblock Lifelong sequential modeling with personalized memorization for user
  response prediction.
\newblock In {\em SIGIR}, pages 565--574, 2019.

\bibitem{repeat}
Pengjie Ren, Zhumin Chen, Jing Li, Zhaochun Ren, Jun Ma, and Maarten de~Rijke.
\newblock Repeatnet: {A} repeat aware neural recommendation machine for
  session-based recommendation.
\newblock In {\em AAAI}, pages 4806--4813, 2019.

\bibitem{fm}
Steffen Rendle.
\newblock Factorization machines.
\newblock In {\em ICDM}, pages 995--1000, 2010.

\bibitem{7927889}
Brent Smith and Greg Linden.
\newblock Two decades of recommender systems at amazon.com.
\newblock {\em IEEE Internet Computing}, 21(3):12--18, 2017.

\bibitem{autoint}
Weiping Song, Chence Shi, Zhiping Xiao, Zhijian Duan, Yewen Xu, Ming Zhang, and
  Jian Tang.
\newblock Autoint: Automatic feature interaction learning via self-attentive
  neural networks.
\newblock In {\em CIKM}, pages 1161--1170, 2019.

\bibitem{bert4rec}
Fei Sun, Jun Liu, Jian Wu, Changhua Pei, Xiao Lin, Wenwu Ou, and Peng Jiang.
\newblock Bert4rec: Sequential recommendation with bidirectional encoder
  representations from transformer.
\newblock In {\em CIKM}, pages 1441--1450, 2019.

\bibitem{caser}
Jiaxi Tang and Ke~Wang.
\newblock Personalized top-n sequential recommendation via convolutional
  sequence embedding.
\newblock In {\em WSDM}, pages 565--573, 2018.

\bibitem{tay2005synthesizer}
Y~Tay, D~Bahri, D~Metzler, DC~Juan, Z~Zhao, and C~Zheng.
\newblock Synthesizer: rethinking self-attention in transformer models (2020).
\newblock {\em arXiv preprint arXiv:2005.00743}.

\bibitem{mlpmixer}
Ilya~O. Tolstikhin, Neil Houlsby, Alexander Kolesnikov, Lucas Beyer, Xiaohua
  Zhai, Thomas Unterthiner, Jessica Yung, Andreas Steiner, Daniel Keysers,
  Jakob Uszkoreit, Mario Lucic, and Alexey Dosovitskiy.
\newblock Mlp-mixer: An all-mlp architecture for vision.
\newblock In {\em NeurIPS}, pages 24261--24272, 2021.

\bibitem{vaswani2017attention}
Ashish Vaswani, Noam Shazeer, Niki Parmar, Jakob Uszkoreit, Llion Jones,
  Aidan~N Gomez, {\L}ukasz Kaiser, and Illia Polosukhin.
\newblock Attention is all you need.
\newblock {\em NeurIPS}, 30, 2017.

\bibitem{escm2}
Hao Wang, Tai{-}Wei Chang, Tianqiao Liu, Jianmin Huang, Zhichao Chen, Chao Yu,
  Ruopeng Li, and Wei Chu.
\newblock {ESCM2:} entire space counterfactual multi-task model for post-click
  conversion rate estimation.
\newblock In {\em SIGIR}, pages 363--372, 2022.

\bibitem{escm2j}
Hao Wang, Zhichao Chen, Jiajun Fan, Yuxin Huang, Weiming Liu, and Xinggao Liu.
\newblock Entire space counterfactual learning: Tuning, analytical properties
  and industrial applications.
\newblock {\em CoRR}, abs/2210.11039, 2022.

\bibitem{mf1}
Haotian Wang, Wenjing Yang, and Naiyang Guan.
\newblock Cauchy sparse {NMF} with manifold regularization: {A} robust method
  for hyperspectral unmixing.
\newblock {\em Knowl. Based Syst.}, 184, 2019.

\bibitem{uplift1}
Haotian Wang, Wenjing Yang, Longqi Yang, Anpeng Wu, Liyang Xu, Jing Ren, Fei
  Wu, and Kun Kuang.
\newblock Estimating individualized causal effect with confounded instruments.
\newblock In {\em {KDD}}, pages 1857--1867. {ACM}, 2022.

\bibitem{wang2020linformer}
Sinong Wang, Belinda~Z Li, Madian Khabsa, Han Fang, and Hao Ma.
\newblock Linformer: Self-attention with linear complexity.
\newblock {\em arXiv preprint arXiv:2006.04768}, 2020.

\bibitem{wang2022image}
Wenhui Wang, Hangbo Bao, Li~Dong, Johan Bjorck, Zhiliang Peng, Qiang Liu, Kriti
  Aggarwal, Owais~Khan Mohammed, Saksham Singhal, Subhojit Som, et~al.
\newblock Image as a foreign language: Beit pretraining for all vision and
  vision-language tasks.
\newblock {\em arXiv preprint arXiv:2208.10442}, 2022.

\bibitem{debias3}
Wenjie Wang, Yang Zhang, Haoxuan Li, Peng Wu, Fuli Feng, and Xiangnan He.
\newblock Causal recommendation: Progresses and future directions.
\newblock In {\em {SIGIR}}, pages 3432--3435. {ACM}, 2023.

\bibitem{wu2017recurrent}
Chao-Yuan Wu, Amr Ahmed, Alex Beutel, Alexander~J Smola, and How Jing.
\newblock Recurrent recommender networks.
\newblock In {\em WSDM}, pages 495--503, 2017.

\bibitem{wu2021fastformer}
Chuhan Wu, Fangzhao Wu, Tao Qi, Yongfeng Huang, and Xing Xie.
\newblock Fastformer: Additive attention can be all you need.
\newblock {\em arXiv preprint arXiv:2108.09084}, 2021.

\bibitem{srgnn}
Shu Wu, Yuyuan Tang, Yanqiao Zhu, Liang Wang, Xing Xie, and Tieniu Tan.
\newblock Session-based recommendation with graph neural networks.
\newblock In {\em AAAI}, volume~33, pages 346--353, 2019.

\bibitem{gcsan}
Chengfeng Xu, Pengpeng Zhao, Yanchi Liu, Victor~S. Sheng, Jiajie Xu, Fuzhen
  Zhuang, Junhua Fang, and Xiaofang Zhou.
\newblock Graph contextualized self-attention network for session-based
  recommendation.
\newblock In {\em IJCAI}, pages 3940--3946, 2019.

\bibitem{ying2018sequential}
Haochao Ying, Fuzhen Zhuang, Fuzheng Zhang, Yanchi Liu, Guandong Xu, Xing Xie,
  Hui Xiong, and Jian Wu.
\newblock Sequential recommender system based on hierarchical attention
  network.
\newblock In {\em IJCAI}, 2018.

\bibitem{yu2022metaformer}
Weihao Yu, Mi~Luo, Pan Zhou, Chenyang Si, Yichen Zhou, Xinchao Wang, Jiashi
  Feng, and Shuicheng Yan.
\newblock Metaformer is actually what you need for vision.
\newblock In {\em CVPR}, pages 10819--10829, 2022.

\bibitem{yuan2020parameter}
Fajie Yuan, Xiangnan He, Alexandros Karatzoglou, and Liguang Zhang.
\newblock Parameter-efficient transfer from sequential behaviors for user
  modeling and recommendation.
\newblock In {\em SIGIR}, pages 1469--1478, 2020.

\bibitem{cnn}
Fajie Yuan, Alexandros Karatzoglou, Ioannis Arapakis, Joemon~M. Jose, and
  Xiangnan He.
\newblock A simple convolutional generative network for next item
  recommendation.
\newblock In {\em WSDM}, pages 582--590, 2019.

\bibitem{zhang2021poolingformer}
Hang Zhang, Yeyun Gong, Yelong Shen, Weisheng Li, Jiancheng Lv, Nan Duan, and
  Weizhu Chen.
\newblock Poolingformer: Long document modeling with pooling attention.
\newblock In {\em ICML}, pages 12437--12446, 2021.

\bibitem{zhang2022efficiently}
Peiyan Zhang, Jiayan Guo, Chaozhuo Li, Yueqi Xie, Jaeboum Kim, Yan Zhang, Xing
  Xie, Haohan Wang, and Sunghun Kim.
\newblock Efficiently leveraging multi-level user intent for session-based
  recommendation via atten-mixer network.
\newblock {\em arXiv preprint arXiv:2206.12781}, 2022.

\bibitem{zhou2019deep}
Guorui Zhou, Na~Mou, Ying Fan, Qi~Pi, Weijie Bian, Chang Zhou, Xiaoqiang Zhu,
  and Kun Gai.
\newblock Deep interest evolution network for click-through rate prediction.
\newblock In {\em AAAI}, volume~33, pages 5941--5948, 2019.

\bibitem{zhou2018deep}
Guorui Zhou, Xiaoqiang Zhu, Chenru Song, Ying Fan, Han Zhu, Xiao Ma, Yanghui
  Yan, Junqi Jin, Han Li, and Kun Gai.
\newblock Deep interest network for click-through rate prediction.
\newblock In {\em SIGKDD}, pages 1059--1068, 2018.

\bibitem{fmlp}
Kun Zhou, Hui Yu, Wayne~Xin Zhao, and Ji{-}Rong Wen.
\newblock Filter-enhanced {MLP} is all you need for sequential recommendation.
\newblock In {\em WWW}, pages 2388--2399, 2022.

\bibitem{zhuograph}
Jianhuan Zhuo, Jianxun Lian, Lanling Xu, Ming Gong, Linjun Shou, Daxin Jiang,
  Xing Xie, and Yinliang Yue.
\newblock Tiger: Transferable interest graph embedding for domain-level
  zero-shot recommendation.
\newblock In {\em CIKM}, page 2806–2816, 2022.

\end{thebibliography}
\bibliographystyle{plain}

\clearpage
\appendix
\renewcommand\thefigure{\Alph{section}\arabic{figure}} 
\setcounter{figure}{0}  
\renewcommand\thetable{\Alph{section}\arabic{table}} 
\setcounter{table}{0} 
\section{Justification of Convolution Theorem}

\begin{lem}\label{lem:a1}
Let $\mathbf{X}=\left[x_1,\dots,x_\mathrm{L}\right]$ and $\mathbf{C}=\left[c_1,\dots,c_\mathrm{L}\right]$ be two $\mathrm{L}$-length sequences. The Fourier transform of a convolution of the two signals is the Hadamard product of their Fourier transforms:
\begin{equation}
    \mathcal{F}(\mathbf{C}*\mathbf{X})=\mathcal{F}(\mathbf{C}) \odot \mathcal{F}(\mathbf{X}).
\end{equation}
\begin{proof}
Assuming that the sequence $\bX$ is periodic with the period $\mathrm{L}$, we have:
\begin{equation}\label{eq:dft_app}
\begin{aligned}
      \mathcal{F}\left(\mathbf{C}*\mathbf{X}\right)_k 
    &\overset{(a)}{=} \sum_{l=0}^{\mathrm{L}-1}\left(\left(\sum_{j=0}^{\mathrm{L}-1} c_j x_{l-j}\right)\exp\left(-\frac{2\pi i}{\mathrm{L}} l k\right)\right) \\
    &\overset{}{=} \sum_{j=0}^{\mathrm{L}-1}c_j\left(\sum_{l=0}^{\mathrm{L}-1} x_{l-j}\exp\left(-\frac{2\pi i}{\mathrm{L}} l k\right)\right) \\
    &\overset{(b)}{=} \sum_{j=0}^{\mathrm{L}-1}c_j\exp\left(-\frac{2\pi i}{\mathrm{L}} j k\right)\left(\sum_{l=0}^{\mathrm{L}-1} x_{l-j}\exp\left(-\frac{2\pi i}{\mathrm{L}} (l-j) k\right)\right) \\
    &\overset{(c)}{=} \left(\sum_{j=0}^{\mathrm{L}-1}c_j\exp\left(-\frac{2\pi i}{\mathrm{L}} j k\right)\right)\left(\sum_{l=0}^{\mathrm{L}-1} x_{l}\exp\left(-\frac{2\pi i}{\mathrm{L}} l k\right)\right) \\
    &\overset{(d)}{=} \mathcal{F}\left(\mathbf{C}\right)_k \odot \mathcal{F}\left(\mathbf{X}\right)_k.
\end{aligned}
\end{equation}

Below is some explanation for the derivation:
\begin{enumerate}[label=(\alph*)]
    \item is the definition of DFT and discrete convolution operation;
    \item breaks down $\exp(2\pi i lk/\mathrm{L})$ into $\exp(2\pi i (l-j)k/\mathrm{L})$ and $\exp(2\pi i jk/\mathrm{L})$;
    \item holds due to the periority of $\bX$ and $\exp(\cdot)$;
    \item is the definition of DFT.
 \end{enumerate}
 
Eq.(\ref{eq:dft_app}) holds for all $0\leq k \leq \mathrm{L}-1$. The proof is completed.
\end{proof}
\end{lem}

\section{Additional experiment results}

\subsection{Performance on the general CTR prediction task}\label{sec:ctr}

In the main text, we have demonstrated the superior performance of ConvFormer in the next-item prediction task using a two-tower retrieval framework. In this section, we aim to highlight the versatility of ConvFormer, specifically the LighTCN layer, as a plug-and-play component that can benefit other general tasks.

We specifically investigate the click-through rate (CTR) estimation task, which involves estimating the CTR and identifying the items that are most likely to be clicked. The CTR estimation task differs from the next-item prediction task in three key aspects: (1) the architecture employed, where CTR estimation typically utilizes a one-tower architecture compared to the two-tower architecture used in next-item prediction; (2) the loss function employed, with CTR estimation using a point-wise loss function as opposed to the pair-wise loss function used in next-item prediction; and (3) the inclusion of user profiles, where CTR estimation involves considering user profiles, unlike next-item prediction which does not require them.
Given the significant differences between CTR prediction and next-item prediction tasks, and considering the wide applications of CTR prediction in practical production scenarios, we choose to include CTR prediction as an additional task to evaluate the performance of ConvFormer.

In CTR estimation frameworks, sequential user models serve as interest extractors, and the resulting user representations are concatenated with user profiles and item properties to estimate the CTR. We implement the interest extractors using the LighTCN layer, as well as comparable methods employed in other baseline models. The performance on the Movie-lens dataset is presented in \autoref{tab:ctr}.
To summarize the main observations:
\begin{itemize}[leftmargin=*]
    \item Replacing the GRU-based interest extractor in DIEN with alternative counterparts shows promise in improving performance in CTR estimation tasks. The relative performance of different models in CTR estimation tasks aligns with their relative performance in next-item prediction tasks.
    \item The original implementation of FMLP-Rec may result in a NAN loss function during training and yield subpar performance. To address this issue, we introduce gradient clipping and enhance the initialization process in FMLP-Rec+. These modifications stabilize the training process, and overall performance surpasses that of SASRec given other settings consistent.
    \item ConvFormer demonstrates the highest overall performance in the CTR estimation task, highlighting the general applicability of the LighTCN layer and the proposed evaluation criteria in various scenarios. Specifically, setting the kernel size as the sequence length (refered to as ConvFormer) without finetuning yields promising results compared to other baseline methods. However, using the full receptive field may not be optimal since early user behaviors could introduce noise and lack informative signals. By fine-tuning the receptive field of the LighTCN layer (referred to as ConvFormer+), further improvements in overall performance can be achieved.
\end{itemize}

\begin{table}
\centering
\caption{Performance comparison as the interest extractor on CTR prediction task. Bold and underlined fonts indicate the first and second best results, respectively.}\label{tab:ctr}
    \setlength{\tabcolsep}{3mm}{
    \small
    \begin{tabular}{lllllll}
    \toprule
    Interest Extractor  &HIT@5 & HIT@10 & HIT@30 & NDCG@5 & NDCG@10 & NDCG@30 \\ \midrule
    GRU4Rec   & 0.4581 & 0.6575 & 0.8887 & 0.2971 & 0.3614 & 0.4173 \\
    SASRec    & 0.4560 & \underline{0.6713} & \underline{0.8940} & 0.2975 & 0.3677 & 0.4216 \\
    FMLP-Rec  & 0.3150 & 0.5546 & 0.8494 & 0.2000 & 0.2772 & 0.3479 \\
    FMLP-Rec+  & 0.4581 & 0.6649 & 0.8897 & \underline{0.3025} & 0.3689 & \underline{0.4236} \\
    \rowcolor{gray!30} ConvFormer& \underline{0.4666} & \textbf{0.6787} & \textbf{0.8982} & 0.3014 & \underline{0.3703} & \underline{0.4236}\\
    \rowcolor{gray!30} ConvFormer+& \textbf{0.5037} & 0.6670 & 0.8929 & \textbf{0.3213} & \textbf{0.3744} & \textbf{0.4290}\\
    \bottomrule
\end{tabular}}
\end{table}

\subsection{Performance of order-sensitive SAR variants}\label{sec:sarvariant}
In \autoref{tab:itemtoitem} we compare the performance of SAR and its variants, illustrating that the simple yet order-sensitive modules can be competitive alternative to the self-attention token-mixer.
We understand that the claims may be aggresive, and it is responsible to ensure the rigor of our experiments. To this end, we have conducted comprehensive experiments on the four benchmarks. We report the results in \autoref{tab:itemtoitem_conv}, as an extension of \autoref{tab:itemtoitem}, showing that the superiority of SAR-O, SAR-P and SAR-R holds across a range of critical hyperparameters (the number of blocks) and random seeds (1-10). We will also open-source the code of these variants, along with the training logs and checkpoint models for each seed, to provide empirical support for our claims and facilitate reproducibility.

The main observations from \autoref{tab:itemtoitem_conv} are summarized as follows.
\begin{itemize}[leftmargin=*]
    \item SAR-O, replacing $\mathbf{A}$ in SAR with a trainable parameter matrix $\mathbf{A}^{(\mathrm{O})}$, outperforms SAR on both benchmarks. The superiority is attributed to the order-sensitivity of the parameter matrix. 
    \item SAR-P, personalizing SAR-O's attention matrix to user behavior patterns, achieves similar performance with SAR-O. The incremental improvement suggest that adaptively generated weights in the item-to-item paradigm are not essential for SAR's leading performance.
    \item SAR-R, fixing SAR-O's weights non-trainable, achieves comparative performance with SAR. Although SAR-R fails to capture semantic relationships between items, it is sensitive to the order of items that is essential to next-item prediction task. As a result, the attention matrix $\mathbf{A}$ in SAR can be replaced with a random matrix $\mathbf{A}^{(\mathrm{R})}$ without performance loss. It is exactly this observation that have inspired our key hypothesis: self-attentive token mixer is not necessarily effective for sequential user modeling. This motivated us to investigate the essences that make Transformer a superior sequential user model, which is one of the major contributions of this work.
    \item SAR-W, without the token mixing process, exhibits a consistent and significant drop in performance compared to SAR. This demonstrates the necessity of token mixing, as even a fixed and randomized token mixer can still maintain competitive performance with SAR.
\end{itemize}

\begin{table}[]
\scriptsize\centering
\setlength{\tabcolsep}{3.5mm}{
\caption{Performance comparison of SAR and variants. Bold fonts indicate the best performance. \textcolor{Maroon}{Red} (resp. \textcolor{PineGreen}{green}) fonts indicate
the variants that are superior (resp. inferior) to  SAR with $p$-value < 0.01 in paired-sample t-test.}\label{tab:variants}
\begin{tabular}{llllllll}
\toprule
\# Layers              & Model & \multicolumn{1}{l}{HIT@1} & \multicolumn{1}{l}{HIT@5} & \multicolumn{1}{l}{HIT@10} & \multicolumn{1}{l}{NDCG@5} & \multicolumn{1}{l}{NDCG@10} & \multicolumn{1}{l}{MRR} \\
\midrule
\multicolumn{8}{c}{\textbf{Beauty dataset}}\\\midrule

\multirow{5}{*}{1} & SAR   & 0.1771                 & 0.3593                 & 0.4512                  & 0.2726                  & 0.3022                   & 0.2736               \\
            & SAR-P & \textbf{0.1778(0.3028)}            & 0.3607(0.1581)            & 0.4519(0.3028)             & 0.2737(0.1920)              & 0.3031(0.2002)              & 0.2746(0.1975)          \\
           & SAR-O & 0.1777(0.2720)             & \textbf{0.3617(0.0554)}            & \textbf{0.4539(0.2720)}              & \textbf{0.2743(0.0603)}             & \textbf{0.3040(0.0165)}               & \textbf{0.2752(0.0515)}          \\
           & SAR-R & 0.1772(0.4342)            & 0.3602(0.2869)            & 0.4521(0.4342)             & 0.2733(0.2821)             & 0.3029(0.2267)              & 0.2743(0.2496)          \\
           & SAR-W & \textcolor{PineGreen}{0.1703(0.9999)}            & \textcolor{PineGreen}{0.3452(1.0000)}                 & \textcolor{PineGreen}{0.4338(0.9999)}             & \textcolor{PineGreen}{0.2622(1.0000)}                  & \textcolor{PineGreen}{0.2907(1.0000)}                   & \textcolor{PineGreen}{0.2641(1.0000)}               \\ \cmidrule{1-8}
\multirow{5}{*}{2} & SAR   & 0.1816                 & 0.3690                  & 0.4631                  & 0.2800                    & 0.3104                   & 0.2804               \\
           & SAR-P & \textbf{0.1820(0.3520)}              & \textbf{0.3696(0.3163)}            & 0.4635(0.3520)              & \textbf{0.2805(0.2861)}             & \textbf{0.3108(0.3262)}              & \textbf{0.2809(0.2848)}          \\
            & SAR-O & 0.1804(0.8409)            & 0.3688(0.5476)            & \textbf{0.4636(0.8409)}             & 0.2791(0.7734)             & 0.3097(0.7387)              & 0.2795(0.7907)          \\
            & SAR-R & 0.1818(0.4456)            & 0.3687(0.6400)              & 0.4614(0.4456)             & 0.2797(0.6307)             & 0.3096(0.7919)              & 0.2801(0.6159)          \\
          & SAR-W & \textcolor{PineGreen}{0.1697(1.0000)}                 & \textcolor{PineGreen}{0.3450(1.0000)}                  & \textcolor{PineGreen}{0.4342(1.0000)}                  & \textcolor{PineGreen}{0.2616(1.0000)}                  & \textcolor{PineGreen}{0.2903(1.0000)}                   & \textcolor{PineGreen}{0.2635(1.0000)}               \\\cmidrule{1-8}
\multirow{5}{*}{3} & SAR   & 0.1822                 & 0.3709                 & 0.4656                  & 0.2810                   & 0.3116                   & 0.2814               \\
            & SAR-P & 0.1830(0.2599)             & 0.3740(0.0308)             & 0.4684(0.2599)             & 0.2831(0.0507)             & 0.3136(0.0598)              & 0.2830(0.0757)           \\
            & SAR-O & \textbf{0.1841(0.0713)}            & \textcolor{Maroon}{\textbf{0.3748(0.0048)}}            & \textbf{0.4695(0.0713)}             & \textcolor{Maroon}{\textbf{0.2841(0.0084)}}             & \textcolor{Maroon}{\textbf{0.3147(0.0061)}}              & \textbf{0.2841(0.0127)}          \\
            & SAR-R & 0.1825(0.4377)            & 0.3724(0.1134)            & 0.4666(0.4377)             & 0.2820(0.2157)              & 0.3125(0.2469)              & 0.2822(0.2712)          \\
            & SAR-W & \textcolor{PineGreen}{0.1726(1.0000)}                 & \textcolor{PineGreen}{0.3477(1.0000)}                 & \textcolor{PineGreen}{0.4379(1.0000)}                  & \textcolor{PineGreen}{0.2644(1.0000)}                  & \textcolor{PineGreen}{0.2935(1.0000)}                   & \textcolor{PineGreen}{0.2665(1.0000)}               \\\cmidrule{1-8}
\multicolumn{8}{c}{\textbf{Sports dataset}}\\\midrule
\multirow{5}{*}{1} & SAR   & 0.1391                 & 0.3316                 & 0.4490                   & 0.2382                  & 0.2761                   & 0.2424               \\
            & SAR-P & 0.1403(0.1274)            & 0.3339(0.0702)            & 0.4526(0.1274)             & 0.2399(0.0881)             & 0.2782(0.0528)              & 0.2440(0.0700)             \\
             & SAR-O & \textbf{0.1416(0.0102)}            & \textbf{0.3344(0.0246)}            & \textbf{0.4528(0.0102)}             & \textcolor{Maroon}{\textbf{0.2409(0.0081)}}             & \textcolor{Maroon}{\textbf{0.2790(0.0045)}}               & \textcolor{Maroon}{\textbf{0.2450(0.0054)}}           \\
            & SAR-R & 0.1410(0.0457)             & 0.3333(0.1171)            & 0.4506(0.0457)             & 0.2399(0.0541)             & 0.2777(0.0481)              & 0.2440(0.0403)           \\
             & SAR-W & \textcolor{PineGreen}{0.1328(0.9997)}            & \textcolor{PineGreen}{0.3148(1.0000)}                 & \textcolor{PineGreen}{0.4275(0.9997)}             & \textcolor{PineGreen}{0.2264(1.0000)}                  & \textcolor{PineGreen}{0.2628(1.0000)}                   & \textcolor{PineGreen}{0.2318(1.0000)}               \\ \cmidrule{1-8}
\multirow{5}{*}{2} & SAR   & 0.1443                 & 0.3442                 & 0.4647                  & 0.2473                  & 0.2861                   & 0.2504               \\
             & SAR-P & \textcolor{Maroon}{\textbf{0.1464(0.0051)}}            & \textcolor{Maroon}{\textbf{0.3480(0.0003)}}             & \textcolor{Maroon}{\textbf{0.4686(0.0051)}}             & \textcolor{Maroon}{\textbf{0.2503(0.0002)}}             & \textcolor{Maroon}{\textbf{0.2891(0.0001)}}              & \textcolor{Maroon}{\textbf{0.2531(0.0003)}}          \\
             & SAR-O & 0.1462(0.0300)              & \textcolor{Maroon}{0.3474(0.0021)}            & 0.4682(0.0300)               & \textcolor{Maroon}{0.2497(0.0059)}             & \textcolor{Maroon}{0.2887(0.0062)}              & \textcolor{Maroon}{0.2526(0.0100)}            \\
             & SAR-R & 0.1446(0.3857)            & 0.3438(0.7097)            & 0.4646(0.3857)             & 0.2470(0.6650)               & 0.2860(0.5882)               & 0.2504(0.5743)          \\
             & SAR-W & \textcolor{PineGreen}{0.1335(1.0000)}                 & \textcolor{PineGreen}{0.3175(1.0000)}                 & \textcolor{PineGreen}{0.4328(1.0000)}                  & \textcolor{PineGreen}{0.2282(1.0000)}                  & \textcolor{PineGreen}{0.2653(1.0000)}                   & \textcolor{PineGreen}{0.2335(1.0000)}               \\\cmidrule{1-8}
\multirow{5}{*}{3} & SAR   & 0.1451                 & 0.3485                 & 0.4709                  & 0.2498                  & 0.2893                   & 0.2528               \\
             & SAR-P & \textcolor{Maroon}{0.1492(0.0042)}            & \textcolor{Maroon}{0.3550(0.0009)}             & \textcolor{Maroon}{0.4792(0.0042)}             & \textcolor{Maroon}{0.2552(0.0021)}             & \textcolor{Maroon}{0.2952(0.0007)}              & \textcolor{Maroon}{0.2576(0.0021)}          \\
             & SAR-O & \textbf{0.1494(0.0127)}            & \textcolor{Maroon}{\textbf{0.3560(0.0004)}}             & \textbf{0.4794(0.0127)}             & \textcolor{Maroon}{\textbf{0.2558(0.0020)}}              & \textcolor{Maroon}{\textbf{0.2956(0.0005)}}             & \textcolor{Maroon}{\textbf{0.2581(0.0025)}}          \\
            & SAR-R & 0.1474(0.0750)             & 0.3531(0.0041)            & 0.4766(0.0750)              & 0.2533(0.0173)             & 0.2931(0.0047)              & 0.2559(0.0189)          \\
             & SAR-W & \textcolor{PineGreen}{0.1339(1.0000)}                 & \textcolor{PineGreen}{0.3177(1.0000)}                 & \textcolor{PineGreen}{0.4326(1.0000)}                  & \textcolor{PineGreen}{0.2285(1.0000)}                  & \textcolor{PineGreen}{0.2655(1.0000)}                  & \textcolor{PineGreen}{0.2338(1.0000)}               \\ \cmidrule{1-8}
\multicolumn{8}{c}{\textbf{Toy dataset}}\\\midrule
\multirow{5}{*}{1} & SAR   & 0.1738                 & 0.3517                 & 0.4475                  & 0.2666                  & 0.2975                   & 0.2690                \\
            & SAR-P & 0.1760(0.0687)             & \textcolor{Maroon}{0.3565(0.0004)}            & \textbf{0.4520(0.0687)}              & \textcolor{Maroon}{0.2701(0.0030)}              & \textcolor{Maroon}{0.3009(0.0048)}              & \textcolor{Maroon}{0.2720(0.0098)}           \\
             & SAR-O & \textcolor{Maroon}{\textbf{0.1765(0.0037)}}            & \textcolor{Maroon}{\textbf{0.3567(0.0000)}}                 & \textcolor{Maroon}{0.4515(0.0037)}             & \textcolor{Maroon}{\textbf{0.2705(0.0000)}}                 & \textcolor{Maroon}{\textbf{0.3011(0.0000)}}                   & \textcolor{Maroon}{\textbf{0.2723(0.0001)}}          \\
             & SAR-R & \textcolor{Maroon}{0.1771(0.0040)}             & \textcolor{Maroon}{0.3542(0.0020)}             & \textcolor{Maroon}{0.4486(0.0040)}              & \textcolor{Maroon}{0.2695(0.0006)}             & \textcolor{Maroon}{0.3000(0.0013)}                 & \textcolor{Maroon}{0.2719(0.0007)}         \\
             & SAR-W & \textcolor{PineGreen}{0.1712(0.9606)}            & \textcolor{PineGreen}{0.3411(1.0000)}                 & \textcolor{PineGreen}{0.4324(0.9606)}             & \textcolor{PineGreen}{0.2598(0.9999)}             & \textcolor{PineGreen}{0.2893(1.0000)}                   & \textcolor{PineGreen}{0.2630(0.9998)}           \\\cmidrule{1-8}
\multirow{5}{*}{2} & SAR   & 0.1790                  & 0.3647                 & 0.4616                  & 0.2759                  & 0.3071                   & 0.2771               \\
             & SAR-P & \textbf{0.1816(0.0383)}            & \textbf{0.3656(0.2032)}            & 0.4620(0.0383)              & \textbf{0.2776(0.0714)}             & \textbf{0.3087(0.0529)}              & \textbf{0.2790(0.0515)}           \\
             & SAR-O & 0.1801(0.1893)            & 0.3650(0.3160)              & \textbf{0.4625(0.1893)}             & 0.2766(0.1591)             & 0.3080(0.1069)               & 0.2780(0.1196)           \\
             & SAR-R & 0.1797(0.2902)            & 0.3634(0.9093)            & 0.4602(0.2902)             & 0.2755(0.6473)             & 0.3068(0.6494)              & 0.2770(0.5346)           \\
             & SAR-W & \textcolor{PineGreen}{0.1697(0.9998)}            & \textcolor{PineGreen}{0.3410(1.0000)}                  & \textcolor{PineGreen}{0.4345(0.9998)}             & \textcolor{PineGreen}{0.2590(1.0000)}                   & \textcolor{PineGreen}{0.2891(1.0000)}                   & \textcolor{PineGreen}{0.2622(1.0000)}               \\\cmidrule{1-8}
\multirow{5}{*}{3} & SAR   & 0.1811                 & 0.3681                 & 0.4661                  & 0.2786                  & 0.3102                   & 0.2796               \\
             & SAR-P & \textbf{0.1830(0.0929)}             & \textbf{0.3703(0.018)}             & \textbf{0.4691(0.0929)}             & \textbf{0.2807(0.0103)}             & \textbf{0.3126(0.0059)}              & \textbf{0.2818(0.0105)}           \\
             & SAR-O & 0.1824(0.2662)            & 0.3698(0.0980)             & 0.4673(0.2662)             & 0.2801(0.1055)             & 0.3115(0.1071)              & 0.2810(0.1538)           \\
             & SAR-R & 0.1792(0.8547)            & 0.3664(0.9009)            & 0.4643(0.8547)             & 0.2769(0.9272)             & 0.3085(0.9502)              & 0.2780(0.9044)           \\
             & SAR-W & \textcolor{PineGreen}{0.1690(1.0000)}                  & \textcolor{PineGreen}{0.3412(1.0000)}                 & \textcolor{PineGreen}{0.4345(1.0000)}                  & \textcolor{PineGreen}{0.2587(1.0000)}                  & \textcolor{PineGreen}{0.2888(1.0000)}                   & \textcolor{PineGreen}{0.2617(1.0000)}               \\\cmidrule{1-8}
\multicolumn{8}{c}{\textbf{Yelp dataset}}\\\midrule
\multirow{5}{*}{1} & SAR   & 0.2199                 & 0.5552                 & 0.7315                  & 0.3922                  & 0.4494                   & 0.3761               \\
             & SAR-P & 0.2208(0.2471)            & 0.5583(0.0285)            & \textbf{0.7332(0.2471)}             & 0.3944(0.0520)              & 0.4511(0.0363)              & 0.3777(0.0723)          \\
             & SAR-O & \textbf{0.2216(0.0698)}            & \textcolor{Maroon}{\textbf{0.5589(0.0095)}}            & 0.7331(0.0698)             & \textbf{0.3950(0.0221)}              & \textbf{0.4515(0.0230)}               & \textbf{0.3783(0.0262)}          \\
             & SAR-R & 0.2199(0.5141)            & 0.5570(0.1078)             & 0.7313(0.5141)             & 0.3931(0.2295)             & 0.4495(0.4106)              & 0.3764(0.3732)          \\
             & SAR-W & \textcolor{PineGreen}{0.2026(1.0000)}                 & \textcolor{PineGreen}{0.5226(1.0000)}                 & \textcolor{PineGreen}{0.6969(1.0000)}                  & \textcolor{PineGreen}{0.3665(1.0000)}                  & \textcolor{PineGreen}{0.4230(1.0000)}                    & \textcolor{PineGreen}{0.3537(1.0000)}               \\\cmidrule{1-8}
\multirow{5}{*}{2} & SAR   & 0.2254                 & 0.5684                 & 0.7446                  & 0.4018                  & 0.4589                   & 0.3842               \\
             & SAR-P & \textcolor{Maroon}{\textbf{0.2290(0.0003)}}             & \textcolor{Maroon}{\textbf{0.5731(0.0000)}}                 & \textcolor{Maroon}{\textbf{0.7473(0.0003)}}             & \textcolor{Maroon}{\textbf{0.4061(0.0000)}}                  & \textcolor{Maroon}{\textbf{0.4626(0.0001)}}              & \textcolor{Maroon}{\textbf{0.3878(0.0001)}}          \\
             & SAR-O & 0.2281(0.0267)            & 0.5713(0.0459)            & 0.7472(0.0267)             & 0.4048(0.0220)              & \textcolor{Maroon}{0.4618(0.0094)}              & \textcolor{Maroon}{0.3870(0.0010)}            \\
            & SAR-R & 0.2275(0.0895)            & 0.5692(0.2404)            & 0.7455(0.0895)             & 0.4033(0.1147)             & 0.4604(0.0861)              & 0.3858(0.0795)          \\
             & SAR-W & \textcolor{PineGreen}{0.2070(1.0000)}                  & \textcolor{PineGreen}{0.5271(1.0000)}                 & \textcolor{PineGreen}{0.7013(1.0000)}                  & \textcolor{PineGreen}{0.3708(1.0000)}                  & \textcolor{PineGreen}{0.4272(1.0000)}                   & \textcolor{PineGreen}{0.3577(1.0000)}               \\\cmidrule{1-8}
\multirow{5}{*}{3} & SAR   & 0.2291                 & 0.5751                 & 0.7498                  & 0.4071                  & 0.4638                   & 0.3886               \\
             & SAR-P & 0.2308(0.1261)            & 0.5787(0.0376)            & \textbf{0.7539(0.1261)}             & 0.4098(0.0720)              & 0.4666(0.0366)              & 0.3908(0.0746)          \\
             & SAR-O & \textcolor{Maroon}{\textbf{0.2324(0.0048)}}            & \textcolor{Maroon}{\textbf{0.5800(0.0038)}}             & \textcolor{Maroon}{0.7538(0.0048)}             & \textcolor{Maroon}{\textbf{0.4115(0.0019)}}             & \textcolor{Maroon}{\textbf{0.4679(0.0004)}}              & \textcolor{Maroon}{\textbf{0.3924(0.0011)}}          \\
             & SAR-R & 0.2314(0.0186)            & 0.5784(0.0332)            & 0.7525(0.0186)             & 0.4098(0.0255)             & 0.4663(0.0118)              & 0.3908(0.0167)          \\
             & SAR-W & \textcolor{PineGreen}{0.2045(1.0000)}                 & \textcolor{PineGreen}{0.5257(1.0000)}                & \textcolor{PineGreen}{0.7014(1.0000)}                  & \textcolor{PineGreen}{0.3693(1.0000)}                  & \textcolor{PineGreen}{0.4262(1.0000)}                   & \textcolor{PineGreen}{0.3564(1.0000)}    \\ 
\bottomrule
\end{tabular}}
\end{table}

Notably, the findings above align with recent literature that challenges the role of self-attention in their respective fields. For example, Google's recent work on abstractive summarization found that replacing the attention matrix with a fixed learnable matrix, such as the SAR-R in our work, could improve most metrics over self-attention (Tab. 3, page 5~\cite{tay2005synthesizer}). They also concluded that The simplest Synthesizers such as Random Synthesizers are fast competitive baseline. (last paragraph, Section 5.2~\cite{tay2005synthesizer}). Similarly, MLP-Mixer~\cite{mlpmixer}, which replaces MHSA with fixed learnable weights, has shown advantages over canonical MHSA in many fundamental and data-rich fields. 

In fact, our findings build up recent DL advances and make reasonable extensions for sequential user modeling. It has been acknowledged that the token-mixer based on non-trainable fixed matrix could achieve promising performance. For example, Goolge concluded that the non-trainable variant performs achieves a strong 24 BLEU with fixed random attention weights~\cite{tay2005synthesizer}; replacing self-attention process with a non-trainable fixed Fourier layer~\cite{lee2022fnet} could achieve 80\% speed-up with a mere 3\%-7\% accuracy drop. Furthermore, we note that SAR-R with a fixed random matrix is more lightweight and sensitive to the order of items than SAR, which are quite important for sequential user modeling. Therefore, it is reasonable to observe that the performance gap between SAR and SAR-R is negligible, with SAR-R even outperforming SAR in some cases.

\section{Reproduction details}
\subsection{Baseline description}

We compare ConvFormer with the following open baselines:
\begin{itemize}[leftmargin=*]
    \item \textbf{PopRec} ranks items based on their popularity, which is determined by the number of user interactions;
    \item \textbf{FM}~\cite{fm} characterizes the pairwise interactions between variables using factorized model;
    \item \textbf{AutoInt}~\cite{autoint} utilizes self-attention mechanism to achieve automatic feature interaction;
    \item \textbf{GRU4Rec}~\cite{gru} encodes user interests with stacked gated recurrent unit;
    \item \textbf{Caser}~\cite{caser} encodes user interests with horizontal and vertical convolution layers;
    \item \textbf{HGN}~\cite{hgn} leverages a hierarchical gating method to model long-short term personalized interest;
    \item \textbf{RepeatNet}~\cite{repeat} strengthens the recurrent neural network with a repetition mechanism which allows selecting items from user behaviors adaptively;
    \item \textbf{CLEA}~\cite{clea} involves an item-level denoising procedure through contrastive learning;
    \item \textbf{SASRec}~\cite{sasrec} is a representative SAR model that utilizes self-attentive token mixer encoders to capture user behavior patterns;
    \item \textbf{BERT4Rec}~\cite{bert4rec} extends SAR with bidirectional encoders and the Cloze training paradigm;
    \item \textbf{SRGNN}~\cite{srgnn} constructs a graph neural network for each session to characterize the transitions of items and enhance their representations;
    \item \textbf{GCSAN}~\cite{gcsan} adds a GNN block to SASRec to capture the local dependencies between nearby items, hence enhancing the potential to learn contextualized item representations.   
    \item \textbf{FMLP-Rec}~\cite{fmlp} replaces the self-attentive token mixer in SASRec with a learnable filter layer, which is the state-of-the-art approach in the context of sequential recommendation.
\end{itemize}
To align the baseline results in related literature, we fully adopt the experimental settings~\cite{fmlp}, including the common hyperparameters (see in \autoref{tab:param}) and training protocol.
We have download the benchmark datasets and baseline source code with the instructions in \cite{fmlp} and have confirmed that results of FMLP-Rec and SASRec are reproducible. 
\subsection{Additional baseline description}

In Section~\ref{sec:empirical}, we design the variants of SAR to examine the key elements of it.
Overall, all models follow the similar paradigm. Specifically, let $\bR\in\mathbb{R}^{\mathrm{L}\times\mathrm{D}}$ be the input representation sequence, these methods generate an attention matrix $\bA$ for fusing the contextual information with in the value vector: $\mathbf{S} = \mathbf{A} (\hat{\mathbf{E}} \mathbf{W}^{(\mathrm{V})})$. The main technical difference among them lies in the generation of $\bA$. See details as follows.
\begin{itemize}[leftmargin=*]
    \item \textbf{SAR} inherits from the standard SASRec approach. It generates the attention matrix$\bA$ through an item-to-item paradigm: 
    \begin{equation}
    \begin{aligned}
        \bQ &= \bR \mathbf{W}^{(\mathrm{Q})} + \bb^{(\mathrm{Q})}, \nonumber \\
        \bK &= \bR \mathbf{W}^{(\mathrm{K})} + \bb^{(\mathrm{K})}, \nonumber \\
        \bA &= \operatorname{softmax}(\bQ\bK^\top\sqrt{\mathrm{D}}) \nonumber.
    \end{aligned}
    \end{equation}
    In order to perform a valid comparison with its alternative counterparts, the multi-head trick is not enabled in this implementation \ie the number of heads is set to 1 as indicated in \autoref{tab:param}. Other components, such as skip connections, dropout, and layer normalizations, remain intact and are kept consistent with the canonical SASRec approach~\cite{sasrec}.
    
    \item \textbf{SAR-O} employs a trainable parameter matrix  $\mathbf{A}^{(\mathrm{O})} \in \mathbb{R}^{\mathrm{L} \times \mathrm{L}}$ as the attention matrix. 
    \item \textbf{SAR-P} utilizes a trainable MLP module to dynamically generate the attention matrix for providing a more tailored experience:
    \begin{equation}
        \bA^{(\mathrm{P})}_l = \bR_l \mathbf{W}^{(\mathrm{P})} + \bb^{(\mathrm{P})},  \quad 0\leq l \leq \mathrm{L}, \nonumber
    \end{equation}
    where $\mathbf{W}^{(\mathrm{P})}\in\mathbb{R}^{\mathrm{D}\times\mathrm{T}}$ and $\bb^{(\mathrm{P})}\in\mathbb{R}^{\mathrm{T}}$ are shared parameters for all time steps.
    \item \textbf{SAR-R} is similar to SAR-O, but its attention matrix $\mathbf{A}^{(\mathrm{R})}$ is randomly initialized, kept fixed, and non-trainable. 
    \item \textbf{SAR-W} is similar to SAR-R, but it has no token-mixer. That is, only the FFN layers are preserved.
    \item \textbf{SAR-N} is similar to SAR, but the query, key and value mappings are not shared at different time steps:
    \begin{equation}
    \begin{aligned}
        \bQ^{(\mathrm{N})}_l &= \bR \mathbf{W}^{(\mathrm{Q})}[l] + \bb^{(\mathrm{Q})}[l],  \quad 0\leq l \leq \mathrm{L}, \nonumber \\
        \bK^{(\mathrm{N})}_l &= \bR \mathbf{W}^{(\mathrm{K})}[l] + \bb^{(\mathrm{K})}[l],  \quad 0\leq l \leq \mathrm{L}, \nonumber \\
        \bA^{(\mathrm{N})} &= \operatorname{softmax}(\bQ^{(\mathrm{N})}\bK^{(\mathrm{N})\top}/\sqrt{\mathrm{D}}) \nonumber.
    \end{aligned}
    \end{equation}
    \item \textbf{SAR-N+} utilizes all items in the input behavior sequence to generate the query, key, and value vectors in SAR. Precisely, let $\mathcal{T}$ be the flatten operator and $\mathcal{T}^{-1}$ be its inverse, the attention matrix is generated by:
    \begin{equation}
    \begin{aligned}
        \bQ^{(\mathrm{N+})} &= \mathcal{T}^{-1}\left(\mathcal{T}(\bR) \mathbf{W}^{(\mathrm{Q+})} + \bb^{(\mathrm{Q+})}\right),  \nonumber \\
        \bK^{(\mathrm{N+})} &= \mathcal{T}^{-1}\left(\mathcal{T}(\bR) \mathbf{W}^{(\mathrm{K+})} + \bb^{(\mathrm{K+})}\right),  \nonumber \\
        \bA^{(\mathrm{N+})} &= \operatorname{softmax}(\bQ^{(\mathrm{N+})}\bK^{(\mathrm{N+})\top}/\sqrt{\mathrm{D}}) \nonumber.
    \end{aligned}
    \end{equation}
    where $\mathcal{T}(\bR)\in\mathbb{R}^{\mathrm{T}\mathrm{D}}$ is the flattened representation sequence, $\mathbf{W}^{(\mathrm{Q+})}\in\mathbb{R}^{\mathrm{T}\mathrm{D}\times \mathrm{T}\mathrm{D}},\mathbf{W}^{(\mathrm{K+})}\in\mathbb{R}^{\mathrm{T}\mathrm{D}\times \mathrm{T}\mathrm{D}}$, $\mathbf{b}^{(\mathrm{Q+})}\in\mathbb{R}^{\mathrm{T}\mathrm{D}},\mathbf{b}^{(\mathrm{K+})}\in\mathbb{R}^{\mathrm{T}\mathrm{D}}$ are the corresponding parameter matrices.
\end{itemize}

In Section~\ref{sec:ablation}, we compare with advanced variants of attentive mechanisms, \ie Fastformer \cite{wu2021fastformer} and PoolingFormer \cite{zhang2021poolingformer}, which outperform a number of effective Transformer variants~\cite{wang2020linformer, beltagy2020longformer}. 
The details of these methods are as follows.
\begin{itemize}[leftmargin=*]
    \item \textbf{PoolingFormer} replaces the self-attentive token mixer with a localized and a large receptive self-attention layer; pooling technology is utilized to the key and value vectors to accelerate the computation of the large receptive self-attention layer.
    \item \textbf{FastFormer} utilizes an additive attention mechanism to model global context, and transform each item's representation based on its interaction with global context representations.
\end{itemize}

We reproduce the Fastformer following its open source implementation at {\url{https://github.com/wuch15/Fastformer}} and the PoolingFormer from scratch.
For the sake of fair comparison, we maintain same values for the common hyperparameters with ConvFormer and SASRec, \eg the learning rate, hidden dimension, number of layers.
Then, we carefully fine-tune their individualized hyperparameters, \eg the pooling size in PoolingFormer, and report their best results. All of these experiments are repeated 11 times with different seeds\footnote{Results are reported with seeds 1-10 and 42.}.

\subsection{Hyperparameter Setting}
\autoref{tab:param} presents the configuration of parameters.
The report of model performance and the finetuning of hyperparameters follow the protocol as follows.
\begin{itemize}[leftmargin=*]
    \item For baselines with publicly available results on the benchmark, we use the results reported based on the work by~\cite{fmlp}. We have checked the results reproduciable and adhered to the same experimental setting. 
    \item For baselines without available results, we reproduce them with careful tuning. For fairness, we maintain the same values for the common hyperparameters with current SOTA baseline and ConvFormer, e.g., the embedding dimension, learning rate, dropout ratio. Then, we finetune their own hyperparameters, e.g., the pooling size in PoolingFormer. 
    \item For SAR-variants in Section~\ref{sec:sarvariant}, these SAR variants rarely introduce new hyperparameters. To make a fair comparison, we use the same common hyperparameters with SAR.
    \item For Convformer, we use the same values of common hyperparameters such as learning rate, number of blocks and dropout rate with the current SOTA baseline by~\cite{fmlp}. Fine-tuning of ConvFormer is limited to its own hyperparameter, K, as detailed in \autoref{tab:param}. Exhaustive tuning of other hyperparams is disabled.
    
\end{itemize}

\begin{table*}[t!]
\centering
\caption{Parameter configurations on each dataset. "*" in the Model field is a wildcard for models.}\label{tab:param}
\setlength{\tabcolsep}{5mm}{
\small
\begin{tabular}{llllll}
\toprule
Parameter & Model & Beauty & Sports & Toys & Yelp \\
\midrule
max sequence length, $\mathrm{L}$ & * & 50 & 50 & 50 & 50 \\
number of layers, $\mathrm{N}$ & * & 2 & 2 & 2 & 2 \\
hidden dimension, $\mathrm{D}$ & * & 64 & 64 & 64 & 64 \\
convolution kernel size, $\mathrm{K}$ & ConvFormer & 45 & 30 & 30 & 30 \\
padding mode & ConvFormer & circular & circular & circular & circular \\
learning rate & * & $1e^{-3}$ & $1e^{-3}$ & $1e^{-3}$ & $1e^{-3}$ \\
weight decay & * & 0 & 0 & 0 & 0 \\
number of attention heads & SAR & 1 & 1 & 1 & 1 \\
number of attention heads & PoolingFormer & 2 & 2 & 2 & 2 \\
number of attention heads & FastFormer & 4 & 2 & 2 & 2 \\
attention dropout probability & * & 0.5 & 0.5 & 0.5 & 0.5 \\
hidden dropout probability & * & 0.5 & 0.5 & 0.5 & 0.5 \\
batch size & * & 256 & 256 & 256 & 256 \\
patience & * & 10 & 10 & 10 & 10 \\
number of maximum epochs & * & 200 & 200 & 200 & 200 \\
pooling size & PoolingFormer & 2 & 4 & 2 & 4 \\
pooling stride & PoolingFormer & 2 & 4 & 2 & 4 \\
local convolution kernel size & PoolingFormer & 10 & 10 & 20 & 20 \\
\bottomrule
	\end{tabular}
	}
\end{table*}
\section{Connecting the dots}
\subsection{Attention v.s. Convolution, a moving average perspective}
The efficacy of model simplification often hinges on precise prior knowledge, prompting an inquiry into why certain simplifications to the Transformer architecture prove effective and what insights they offer. The self-attentive architecture can be conceptualized as a moving average (MA) model in the value sequence $\bV$ as $\bS = \bA \bV$, where the weights $\bA$ are dynamically generated from the inputs. While the dynamic weight augments model capacity, it  encounters limitations in sequential user modeling: (1) order-sensitivity; (2) instability, where the model parameters vary with the input, leading to difficulties in parameter identification.
SAR-O, which introduces an input-independent parameter matrix, effectively mitigates these limitations and enhances performance. This suggests that traditional MA models retain significant research and practical value in this domain.

The MA model is designed for Markov process identification problems. Markov processes have two important properties: (1) order, i.e. the number of the previous steps related to the current state, and (2) coupling, i.e. whether the update of a particular channel depends on the states in other channels. Our findings indicate that increasing the receptive field and minimizing channel coupling improve performance, suggesting that user behavior sequences in latent space exhibit high-order, decoupled Markovian properties. Techniques like optimal filtering and smoothing offer robust inference methods for such processes, as exemplified by the success of Kalman attention \cite{liu2020kalman}, which opens up promising avenues for future research.

\subsection{Large receptive field v.s. lightweight architecture, a learning theory perspective}
In Criteria 2 and 3, we advocate for a large receptive field coupled with a lightweight architecture as essential elements for effective token mixers. While our experimental results substantiate these criteria, we further elucidate their theoretical underpinnings through the lens of Probably Approximately Correct (PAC) learning, based on Lemma~\ref{lem:2}. 
\begin{lem}
\label{lem:2}
Let $g$ be the hypothesis (model) selected by the learning algorithm with the `statistical' large dataset $\mathcal{D}$. Let $E_\mathrm{in}(g)$ and $E_\mathrm{out}(g)$ be the within-sample error and out-of-sample error of the selected model $g$, the generalization gap is defined as
\begin{equation}
    \delta(g):=E_\mathrm{out}(g) - E_\mathrm{in}(g).
\end{equation}
If $\delta$ is significantly large, overfitting happens and learning fails. Furthermore, $g$ can be bounded as:
\begin{equation}
P_{\mathcal{D}}[\underbrace{\left|E_\mathrm{out}(g)-E_\mathrm{in}(g)\right|}_{\delta(g)}>\mathrm{\epsilon}]\leq4(2\mathrm{M})^{\mathrm{d_{vc}}} \exp \left(-\frac{1}{8} \mathrm{\epsilon}^{2} \mathrm{M}\right),
\end{equation}
where $\mathrm{M}$ is the sample size of the dataset $\mathcal{D}$, 
$\mathrm{d_{vc}}$ is the VC-dimension that measures the complexity of the model, $\epsilon$ is a confidence threshold.

\end{lem}

Employing a large receptive field enables efficient capture of long-term patterns, offering advantage over stacking multiple small kernels which can distort information through successive non-linear transformations, thus enabling a reduced within-sample error $E_\mathrm{in}$. However, this approach increases model complexity measured by $\mathrm{d_{vc}}$, thereby widening the generalization gap $\delta$. To mitigate this, a lightweight architecture, achieved through techniques like parameter sharing or inductive bias, is necessary. Both strategies cooperate to effectively minimize training error and control the generalization gap, consequently reducing the generalization error $E_\mathrm{out}$.

\section{More comparison with existing methods}
ConvFormer is a non-trivial advancement in the field of sequential user modeling albeit with a relatively simple architecture. This is evidenced by the observation that it is not feasible to obtain ConvFormer by simply adding a simple update to existing sequential user models. The details are formulated as follows.
\subsection{Comparison with current CNN-based solutions}
There are several major differences between ConvFormer and the conventional convolution networks, particularly Caser~\cite{caser}, an exemplar CNN-based sequential user model. ConvFormer incorporates a meta-former architecture with a residual link, layer normalization, and disentanglement between the token-mixer and the channel-mixer, which makes it fundamentally different from canonical convolution networks\footnote{The incorporation of meta-former architecture is a non-trivial technical point that has recently received attention from the machine learning community~\cite{yu2022metaformer}.}. 
Besides, the DWC layer in ConvFormer is fundamentally different from the horizontal and vertical convolution layers~\cite{caser} which are commonly used in sequential user models. The technical differences and advantages of the DWC layer are illustrated in \autoref{tab:diff1}. To summarize:
\begin{itemize}[leftmargin=*]
    \item The horizontal convolution~\cite{caser} is essentially a canonical convolution followed by a max-pooling layer, and even with a large receptive field, it is not equivalent to our DWC layer. In fact, the horizontal convolution layer is similar to the Conv-V variant in Section~\ref{sec:lightconv}. According to \autoref{fig:paramshared_conv}, the Conv-V variant is obviously inferior to the DWC layer, which is a reasonable result since it violates our criteria of lightweight architecture. 
    \item The vertical convolution is actually a canonical multi-kernel 2D convolution with kernel shape $\mathrm{1}\times\mathrm{L}$, cascaded by a max-pooling layer. In contrast, the DWC layer is a depth-wise single-kernel 2D convolution with kernel shape $\mathrm{D}\times\mathrm{L}$, free of a cascaded pooling layer.
\end{itemize}

\begin{table}[]
\caption{DWC vs convolution-based sequential user models~\cite{caser}.}\label{tab:diff1}
\tiny
\centering
\begin{tabular}{p{2.5cm}p{2cm}p{2cm}p{2cm}p{5cm}}
\toprule
& DWC layer              & Horizontal convolution & Vertical convolution  & Advantage of the DWC layer     
                                                                       \\
                                                                       \midrule
\# convolution kernel                                                                & 1                      & Z                      & Z                     & The DWC layer is more lightweight                                                                                                                                                                                                    \\\midrule
Convolution kernel                                                               & depth-wise convolution & canonical convolution  & canonical convolution & The DWC layer is more lightweight                                                                                                                                                                                                    \\\midrule
\begin{tabular}[c]{@{}l@{}}Number of parameters\\ (given full receptive field)\end{tabular} & $\mathrm{D}\times \mathrm{L}$                    & $\mathrm{D}\times \mathrm{L}\times \mathrm{Z}$                  & $\mathrm{L}\times \mathrm{Z}$                     & The DWC layer avoids an extra hyper-parameter Z                                                                                                                                                                                      \\\midrule
Pooling-demanding                                                                           & \XSolidBrush                      & \Checkmark                      & \Checkmark                     & The DWC layer does not incorporate pooling and preserves the ordering information                                                                                                                                                    \\\midrule
Padding                                                                                     & \Checkmark                      & \XSolidBrush                      & \XSolidBrush                     & The DWC layer incorporates a padding operation to ensure its input and output have the same shape, enabling residual link                                                                                                            \\\midrule
Receptive field                                                                             & large                  & limited                & L                     & The DWC layer controls the risk of overfitting and thus makes it feasible to incorporate large receptive field                                                                                                                       \\\midrule
Meta-former architecture                                                                    & \Checkmark                      & \XSolidBrush                      & \XSolidBrush                     & The DWC layer adapts the meta-former architecture. Notably, given meta-former architecture, the DWC layer remains superior than other token-mixers, see section 5.3 for reference.                                                   \\\midrule
Complexity (accelerated)                                                                    & $\mathcal{O}(\mathrm{D}\times \mathrm{L}\log \mathrm{L})$            & $\mathcal{O}(\mathrm{Z}\times \mathrm{L}\log \mathrm{L}\times \mathrm{D} \log \mathrm{D})$     & $\mathcal{O}(\mathrm{Z}\mathrm{D}\times \mathrm{L} \log \mathrm{L})$         & The DWC layer performs 1-D convolution in each channel, thus can be accelerated with 1D  FFT; the horizontal convolution performs 2-D convolution and employs Z kernels, thus can be accelerated with 2D FFT with larger complexity.\\
\bottomrule
\end{tabular}
\end{table}
\subsection{Comparison with MLPs}
A potential concern might arise regarding the Depth-Wise Convolution (DWC) layer reducing to a Multi-Layer Perceptron (MLP) under a full receptive field (FRF) setting. However, key differences exist between the two, as outlined in \autoref{tab:diff2}. To highlight the core differences:
\begin{itemize}[leftmargin=*]
    \item Acceleration: The Fast Fourier Transform (FFT) can expedite computations of DWC layers, a feature not applicable to MLPs. This underscores a fundamental difference in computational efficiency.
    \item Weight Dimensions: For a one-dimensional input sequence $X\in\mathbb{R}^{\mathrm{L}\times1}$, an MLP would require a weight matrix $W\in \mathbb{R}^{\mathrm{L}\times \mathrm{L}}$ to maintain output shape (for residual link). In contrast, a DWC layer with FRF would only need $W\in \mathbb{R}^{\mathrm{K}\times 1}$ with $\mathrm{K}=\mathrm{L}$, highlighting its lightweight nature even in the FRF setting.
\end{itemize}

\begin{table}[]
\tiny
\centering
\caption{MLP vs  DWC with Full Receptive Field.}\label{tab:diff2}
\begin{tabular}{p{2.5cm}p{3.5cm}p{3.5cm}p{5cm}}
\toprule
Technical difference     & DWC layer (FRF)                                    & MLP                                                                                     & Advantage of the DWC layer                                                                                    \\ \midrule
Number of parameters     & $\mathrm{D}\times\mathrm{L}$ because each hidden dimension has unique convolution weight. & $\mathrm{L}\times\mathrm{L}\times\mathrm{D}$ with unique MLP per hidden dimension, $\mathrm{L}\times\mathrm{L}$ with shared MLP across dimensions. & The number of parameters in a DWC layer is comparatively lower than that of an MLP, particularly when dealing with longer sequences.                                \\\midrule
Meta-former architecture & \Checkmark                                                  & \XSolidBrush                                                                                       & The DWC layer is based on the Meta-Former architecture, which distinguishes it from canonical MLPs.                      \\\midrule
Accelerable with FFT     & \Checkmark                                                  & \XSolidBrush                                                                                       & The FFT can accelerate the computation of the DWC layer, but it does not speed up the computation of MLP.\\
\bottomrule
\end{tabular}
\end{table}

\section{Broader Impact}
ConvFormer serves as a proof-of-concept for the proposed criteria, yielding superior performance over various prevalent solutions in sequential user modeling. To further demonstrate the versatility of our proposed criteria, we introduce L-Mixer. The difference with Transformer and ConvFormer is that it uses an affine layer acting as the token mixer, which is embarassingly simple yet satisties the proposed criteria simultaneously. The results are summarized in \autoref{tab:main_table_mlp}. While L-Mixer shows promise, it lacks adaptability to varying input lengths during inference and cannot be accelerated like the LighTCN layer. Hence, ConvFormer remains our primary example in the main text. We hope that the success of these two simple yet effective modifications could provide inspirations to design effective token mixers in sequential user models.

\begin{table}[t!]
\scriptsize\centering
	\caption{Performance on four datasets. Bold and underlined fonts indicate the best and second-best result, respectively. "*" marks the significant improvement over the second-best result with p-value $<$ 0.01 on the one-sample t-test.}
	\label{tab:main_table_mlp}
	\setlength{\tabcolsep}{2.2mm}{   
	\begin{tabular}{llccccccccc}
	\toprule
    Dataset & Metric &FM& AutoInt &GRU4Rec & Caser &SASRec &BERT4Rec &GCSAN &FMLP &L-Mixer  \\
	\midrule
\multirow{6} * {Beauty}
 &H@1   &0.0405 &0.0447 &0.1337 &0.1337  &0.1870 &0.1531 &{0.1973} &\underline{0.2011} &\textbf{0.2020}\\
 &H@5   &0.1461 &0.1705 &0.3125 &0.3032  &0.3741 &0.3640 &0.3678   &\underline{0.4025}  &{\textbf{0.4151}}\sig\\
 &N@5   &0.0934 &0.1063 &0.2268 &0.2219  &0.2848 &0.2622 &0.2864   &\underline{0.3070}  &{\textbf{0.3143}}\sig\\
 &H@10  &0.2311 &0.2872 &0.4106 &0.3942  &0.4696 &0.4739 &0.4542   &\underline{0.4998}  &{\textbf{0.5139}}\sig\\
 &N@10  &0.1207 &0.1440 &0.2584 &0.2512  &0.3156 &0.2975 &0.3143   &\underline{0.3385}  &{\textbf{0.3462}}\sig\\
 &MRR   &0.1096 &0.1226 &0.2308 &0.2263  &0.2852 &0.2614 &0.2882   &\underline{0.3051}  &{\textbf{0.3105}}\sig\\
\midrule
\multirow{6} * {Sports}
 &H@1   &0.0489 &0.0644 &0.1160 &0.1135  &0.1455 &0.1255 &\underline{0.1669} &{0.1646} &\textbf{0.1652}\\
 &H@5   &0.1603 &0.1982 &0.3055 &0.2866  &0.3466 &0.3375 &{0.3588} &\underline{0.3803} &\textbf{0.3919}\sig\\
 &N@5   &0.1048 &0.1316 &0.2126 &0.2020  &0.2497 &0.2341 &{0.2658} &\underline{0.2760} &\textbf{0.2823}\sig\\
 &H@10  &0.2491 &0.2967 &0.4299 &0.4014  &0.4622 &0.4722 &{0.4737} &\underline{0.5059} &\textbf{0.5150}\sig\\
 &N@10  &0.1334 &0.1633 &0.2527 &0.2390  &0.2869 &0.2775 &{0.3029} &\underline{0.3165} &\textbf{0.3221}\sig\\
 &MRR   &0.1202 &0.1435 &0.2191 &0.2100  &0.2520 &0.2378 &{0.2691} &\underline{0.2763} &\textbf{0.2804}\sig\\
\midrule
\multirow{6} * {Toys}
 &H@1   &0.0257 &0.0448 &0.0997 &0.1114  &0.1878 &0.1262 &\underline{0.1996} &{0.1935} &\textbf{0.1984}\\
 &H@5   &0.0978 &0.1471 &0.2795 &0.2614  &{0.3682} &0.3344&0.3613 &\textbf{0.4063} &\underline{0.4052}\\
 &N@5   &0.0614 &0.0960 &0.1919 &0.1885  &0.2820 &0.2327 &{0.2836} &\underline{0.3046} &\textbf{0.3068}\\
 &H@10  &0.1715 &0.2369 &0.3896 &0.3540  &{0.4663} &0.4493 &0.4509 &\underline{0.5062} &\textbf{0.5053}\\
 &N@10  &0.0850 &0.1248 &0.2274 &0.2183  &{0.3136} &0.2698 &0.3125 &\underline{0.3368} &\textbf{0.3391}\\
 &MRR   &0.0819 &0.1131 &0.1973 &0.1967  &0.2842 &0.2338 &0.{2871} &\underline{0.3012} &\textbf{0.3043}\\
\midrule
\multirow{6} * {Yelp}
 &H@1   &0.0624 &0.0731 &0.2053 &0.2188  &0.2375 &0.2405   &{0.2493} &\underline{0.2727} &\textbf{0.2797}\\
 &H@5   &0.2036 &0.2249 &0.5437 &0.5111  &0.5745 &{0.5976} &0.5725 &\underline{0.6191} &\textbf{0.6260}\sig\\
 &N@5   &0.1333 &0.1501 &0.3784 &0.3696  &0.4113 &{0.4252} &0.4162 &\underline{0.4527} &\textbf{0.4602}\sig\\
 &H@10  &0.3153 &0.3367 &0.7265 &0.6661  &0.7373 &{0.7597} &0.7371 &\underline{0.7720} &\textbf{0.7721}\\
 &N@10  &0.1692 &0.1860 &0.4375 &0.4198  &0.4642 &{0.4778} &0.4696 &\underline{0.5024} &\textbf{0.5077}\sig\\
 &MRR   &0.1470 &0.1616 &0.3630 &0.3595  &0.3927 &{0.4026} &0.4006 &\underline{0.4299} &\textbf{0.4360}\sig\\
\bottomrule
\end{tabular}
}
\end{table}

\end{document}